\documentclass[10pt,twocolumn,letterpaper]{article}
\usepackage[pagenumbers]{wacv} 
\usepackage{tikz}
\usepackage{graphicx}
\usepackage{float}
\usepackage{amsmath}
\usepackage{amssymb}
\usepackage{mathtools}
\usepackage{booktabs}
\usepackage{multirow}
\usepackage{makecell}
\usepackage{pifont}
\usepackage{tocloft}
\usepackage{tcolorbox}
\usepackage[flushleft]{threeparttable}
\usepackage[table]{xcolor} 
\usepackage{wrapfig}
\usepackage[linesnumbered, commentsnumbered, ruled, vlined]{algorithm2e}

\usepackage{xcolor}
\newcommand*\colourcheck[1]{%
  \expandafter\newcommand\csname #1check\endcsname{\textcolor{#1}{\ding{52}}}%
}
\colourcheck{blue}
\colourcheck{green}
\colourcheck{red}
\definecolor{mygray}{gray}{.9}
\definecolor{mylgray}{gray}{.96}

\newcommand\crossmark[1][]{%
  \tikz[scale=0.4,#1]{
    \fill(0,0)--(0.1,0) .. controls (0.5,0.4) .. (1,0.7)--(0.9,0.7) ..  controls (0.5,0.5) ..(0,0.1) --cycle;
    \fill(1,0.1)--(0.9,0.1) .. controls (0.5,0.3) .. (0,0.7)--(0.1,0.7) .. controls (0.5,0.4) ..(1,0.2) --cycle;
  }%
}

\usepackage{subcaption}

\usepackage{colortbl}

\usepackage{float}
\usepackage{setspace}
\usepackage{multirow}
\usepackage{makecell}
\usepackage{breqn}

\AtEndDocument{
  \addtocontents{toc}{\protect\par\noindent\rule{\linewidth}{1pt}}%
}
\usepackage{listings}
\usepackage{etex}  
\definecolor{wacvblue}{rgb}{0.21,0.49,0.74}
\usepackage[pagebackref,breaklinks,colorlinks,allcolors=wacvblue]{hyperref}

\def\wacvPaperID{617} 
\def\confName{WACV}
\def\confYear{2026}

\title{{FedSCAl}: Leveraging Server and Client Alignment for Unsupervised Federated Source-Free Domain Adaptation}

\author{M. Yashwanth$^{1}$$^{*}$, Sampath Koti$^{1}$$^{*}$, Arunabh Singh$^{1}$$^{*}$, Shyam Marjit$^{1}$$^{*}$, and Anirban Chakraborty$^{1}$\\
$^{1}$ Indian Institute of Science, Bengaluru, India\\
 {\tt\small{\{yashwanthm,sampathkoti,shyammarjit,anirban\}@iisc.ac.in}},
\tt\small{arunabhsingh25@gmail.com}}

\begin{document}

\twocolumn[{
\renewcommand\twocolumn[1][]{#1}%
\maketitle
\begin{center}
\vspace{-6mm}
Project Page: \href{https://vcl-iisc.github.io/FedSCAl/}{https://vcl-iisc.github.io/FedSCAl/}
\end{center}
}]

\def\thefootnote{*}\footnotetext{Equal contribution.}

\vspace{-5.0in}
\begin{abstract}

We address the Federated source-Free Domain Adaptation (FFreeDA) problem, with clients holding unlabeled data with significant inter-client domain gaps. The FFreeDA setup constrains the FL frameworks to employ only a pre-trained server model as the setup restricts access to the source dataset during the training rounds. Often, this source domain dataset has a distinct distribution to the clients' domains. 
To address the challenges posed by the FFreeDA setup, adaptation of the Source-Free Domain Adaptation (SFDA) methods to FL struggles with client-drift in real-world scenarios due to extreme data heterogeneity caused by the aforementioned domain gaps, resulting in unreliable pseudo-labels. In this paper, we introduce FedSCAl, an FL framework leveraging our proposed Server-Client Alignment (SCAl) mechanism to regularize client updates by aligning the clients' and server model's predictions. We observe an improvement in the clients' pseudo-labeling accuracy post alignment, as the SCAl mechanism helps to mitigate the client-drift. Further, we present extensive experiments on benchmark vision datasets showcasing how FedSCAl consistently outperforms state-of-the-art FL methods in the FFreeDA setup for classification tasks. 
\end{abstract}    
\vspace{-0.2in}
\section{Introduction}
\label{sec:intro}
\vspace{-0.04in}
Federated Learning (FL) initially proposed in~\cite{mcmahan2017communication} is a machine learning paradigm where multiple client models actively participate in the collective training of a shared global model, orchestrated by a central server. During the training phase, clients exchange their local client models with the server, thus ensuring the confidentiality of their respective local training data. The privacy-preserving nature of FL has led to its application across diverse domains, including smartphones~\cite{47586,ramaswamy2019federated}, Internet of Things~\cite{mills2019communication,nguyen2021federated}, and healthcare applications~\cite{rieke2020future,xu2021federated,nguyen2022federated,brisimi2018federated}. 


\begin{figure}[!ht]
    \centering
        \includegraphics[scale=1.8]{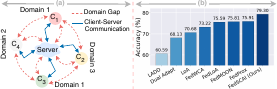}
        \vspace*{-4mm}
        \caption{(a) \textit{FFreeDA Setup:} \textit{ Multiple clients (labeled as $\text{C}_1$, $\text{C}_2$, $\text{C}_3$, and $\text{C}_4$) hold unlabeled data from same or different domains, while the central server uses a pre-trained model trained on a labeled source dataset, which differs from clients' data distributions and is unavailable during training. (b) Performance of different methods on Office-Home dataset.}}
        \label{teaser_plot1}
        \vspace{-5mm}
\end{figure}

In practice, clients participating in an FL framework will be deployed in diverse environments, ingesting non-iid data leading to data heterogeneity across the client devices. Heterogeneity in data often arises due to variations in label distributions or feature distributions (covariate shift) across the clients. In their respective local training rounds, the clients optimize a task-specific objective function on their training data. Consequently, clients may drift towards the local minimum of their respective objectives. This phenomenon, `client-drift'~\cite{karimireddy2020scaffold} is commonly observed when training data is non-iid across clients. As a result, client models tend to overfit their training data. Furthermore, aggregating these model parameters at the server results in a sub-optimal global model.

Prior works in FL have addressed label heterogeneity issues in ~\cite{karimireddy2020scaffold, gao2022feddc, li2020federated, li2021model, Yashwanth_2024_WACV} while covariate shift has been considered in~\cite{Huang_2023_CVPR,li2021fedbn}. 
These methods assume that the client data is labeled, which is often impractical given the data privacy as well as logistical constraints associated with data annotation. Evidently, in practical real-world scenarios (as mentioned in Sec. 2.2 of the Supplementary), FL setups usually encounter substantial challenges with unlabeled client data and large domain gaps across the clients 
hindering the development of a robust aggregated model that generalizes well across diverse client environments. In this work, we take up the classification task in a challenging FL setting of FFreeDA introduced in~\cite{shenaj2023learning}. In this setting, the training data samples on the clients are unlabeled and, the server model is pre-trained on the source dataset. This dataset is unavailable during the federated training, and the FL is initialized by the pre-trained model available on the server. Along with inter-client domain shift, the source dataset used for pre-training the server model is likely to exhibit a large domain gap with each of the clients' training data. This setting is clearly explained in Figure~\ref{teaser_plot1} (a).

Existing methods that require labeled source data on the server, such as DualAdapt~\cite{yao2022federated}, are not effective in the FFreeDA setting, where the server lacks access to labeled data. Even when access is permitted, these methods often fail to bridge significant domain gaps. More recent approaches such as LADD~\cite{shenaj2023learning} for segmentation and FedWCA~\cite{mori2025federated} for classification adopt clustering-based strategies tailored to the FFreeDA paradigm. However, their performance is dependant on the clustering quality, making them vulnerable when the cluster structure is unclear. In particular, LADD’s requirement for client-style information introduces privacy concerns that limit its applicability. 

Given the unsupervised nature of FFreeDA, we turn to pseudo-labeling approaches based on source-free domain adaptation, such as SHOT~\cite{liang2020we}, that allow each client to self-train using predictions from the current model. We begin with two simple baselines: Local Adaptation (LoA), where each client trains independently using pseudo-labels, and its federated counterpart FedLoA, which aggregates client models after local adaptation. As shown in Figure~\ref{teaser_plot1}~(b), both methods perform poorly when the client domains are highly non-overlapping, due to inconsistent label assignments and client drift.

To overcome these issues, we propose FedSCAl, a new framework that replaces clustering with a Server and Client Alignment (SCAl) mechanism. Rather than grouping clients, FedSCAl aligns client predictions with those from a server model. This alignment leads to more stable and accurate pseudo-labels, thereby mitigating client drift and improving generalization across diverse domains.

Additionally, we propose a novel adaptive thresholding approach within our framework that adjusts the threshold used for determining the pseudo-labels dynamically based on the entropy of the predictions on training data samples.
This further enhances the quality of pseudo labels used for training the client models across the training rounds. We discuss this in detail in Sec. \ref{adapt_thr}.

Our key contributions are summarised as follows:

\begin{itemize}   
 \item  \textbf{FedSCAl 
 framework}:     A novel framework to solve classification tasks in the FFreeDA~\cite{shenaj2023learning} setting where only a pre-trained source model is available at the server, and clients hold unlabeled, heterogeneous data.

 \item  \textbf{SCAl mechanism}: A server–client alignment mechanism within the FedSCAl framework that regularizes client predictions using the server model, mitigating client-drift and improving pseudo-label accuracy.
 \
  \item \textbf{Adaptive Thresholding}: A dynamic approach for filtering low-confidence predictions, ensuring the reliability of pseudo-labels used for client updates.

  \item \textbf{Extensive Empirical Validation}: Experiments on multiple benchmarks (Office-Home, DomainNet, and Office-31) demonstrate substantial performance gains over the baselines and other state-of-the-art approaches.
\end{itemize}

\section{Related Work}
\label{sec: Related_Work}
\subsection{Federated Learning (FL)}
Recently, addressing non-iid issues in Federated Learning has become a rapidly evolving research area. Here, we briefly discuss a few related works. In FedAvg~\cite{mcmahan2017communication}, the two main challenges explored are reducing communication costs~\cite{yadav2016review} and ensuring privacy by avoiding having to share the data. 
FedProx~\cite{li2020federated} introduced a proximal term by penalizing the weights if they are far from the global initialized model. 
SCAFFOLD~\cite{karimireddy2020scaffold} approached this issue by considering it as an objective inconsistency, and it introduced a gradient correction term designed to function as a regularizer.
Later works of~\cite{gao2022feddc,karimireddy2020scaffold} improved upon this by introducing the dynamic regularization term. 
A few works include regularization methods on the client side~\cite{zhu2021data}, and one-shot methods where clients send the condensed data and the server trains on the condensed data~\cite{zhou2020distilled}. 
In~\cite{hsu2020federated}, an adaptive weighting scheme is considered on task-specific loss to minimize the learning from samples whose representation is negligible. Flatness-based methods based on SAM called FedSAM is introduced in~\cite{qu2022generalized,caldarola2022improving,sun2023fedspeed}. Federated Learning with partially labelled data is considered in the works of SemiFL~\cite{diao2022semifl,jeong2021federated}. Model personalizations in FL are considered in~\cite{tan2022towards}
and the distillation-based approaches are also considered in~\cite{he2022class,lee2021preservation}. The clustering approaches in FL are discussed in~\cite{ghosh2020efficient,long2023multi} where they require retraining the models repeatedly to obtain optimal clustering in a supervised learning setup.
\subsection{Federated Learning with Distributional Shifts}
FL has seen few developments concerning distributional shifts. The work of~\cite{peng2019federated} introduces FL with distributional shifts, framing the challenge as adversarial domain adaptation within FL settings. Personalized FL for labelled data with distributional shifts is explored in~\cite{li2021fedbn}. FedGen~\cite{seo202216} adopts a distillation approach, training generators on the client side to minimize the domain gap across clients. 
However, existing approaches, as outlined in~\cite{ li2021fedbn, Huang_2023_CVPR, yao2022federated}, share common limitations. They rely on the assumption of labelled data across clients or the availability of related domain data with the server~\cite{yao2022federated}. Methods such as DualAdapt~\cite{yao2022federated} depends on a labelled source dataset during federated training, and LADD~\cite{shenaj2023learning} assumes access to client data styles in the pre-training stage. Few prior works such as Semifl~\cite{diao2022semifl} use consistency regularization inspired from Fixmatch~\cite{sohn2020fixmatch}, where the weak and strong augmentations are used to produce better pseudo-labels. 

\section{Preliminary}
\label{sec:Pre_FL}
\subsection{Pseudo-Label Estimation}
\label{Pre:sfda}
In the FFreeDA setting, 
access to the labelled source domain data $D^s$ is not available. Instead, the federation relies on a pre-trained source model $f^s$   trained on dataset  $D^s$ to facilitate adaptation to the target domain $D^t$.
In such settings, the principal objective is to learn a target model $f^t$ which could correctly predict labels $y_t$ for the unlabelled target domain samples $x_t\in D^t$ by adaptation from the source-trained model $f^s$ without direct access to any labelled data from the source domain $D^s$. The challenge lies in leveraging the knowledge encoded within the pre-trained source model $f^s$ to adapt effectively to the characteristics of the target domain $D^t$
.
The source model $f^s$ can be studied as a composition of two integral components: a feature extractor $g^s: X_s \rightarrow \mathbb{R}^d$, where $X_s$ is the feature distribution of source data, and a classifier $h^s: \mathbb{R}^d \rightarrow \mathbb{R}^J$. This architecture enables the transformation of input data $x$ from the source domain to a $d$-dimensional feature space, followed by classification into $J$ classes.
In other words, the source model $f^s$ can be represented as $f^s ={ h^s \circ g^s}$. As a result of the domain gap between the server and client models we employ a pseudo-label estimation strategy as in ~\cite{liang2020we} which helps in knowledge transfer via feature prototype-based pseudo-labeling strategy, where prototypes $c_j$  are obtained for each class akin to weighted k-means clustering, computed as:=
\vspace{-0.1in}
\begin{equation}
c_{j} = \frac {\sum_{x_t\in D^t} \delta _j({f^t}(x_t))\:  {g^t}(x_t)}{\sum_{x_t\in D^t}\delta _j( {f^t}(x_t))},
\end{equation}
\vspace{-0.02in}
where ${f^t} = {h^t} \circ {g^t}$ denotes the target model, $\delta_j({f^t}(x^i_t))$ represents the softmax probability of target instance $x^i_t$ belonging to the $j$-th class.
The pseudo label $\hat{y}^i_t$ for target $x^i_t$ is determined via the nearest prototype classifier: $\hat {y}^i_t = \arg \min _j D_f({g^t}(x^i_t), c_j)$
where $D_f(a, b)$ computes the cosine distance between $a$ and $b$. Therefore, given the source model \( f^s = h^s \circ g^s \) and pseudo labels generated as above, SHOT freezes the hypothesis from the source \( h^t = h^s \) and learns the feature encoder \( g^t \) with the full objective Eq.~\ref{eq_shot}.  It jointly minimizes the Information Maximization (IM) loss ($L_{\text{ent}} + L_{\text{div}}$)~\cite{hu2017learning,shi2012information} together with $L_{pce}$, which is the pseudo-labeling loss with the full objective as
\begin{equation}
    L^{LoA} = L_{\text{ent}} + L_{\text{div}} + L_{pce} 
\label{eq_shot}
\end{equation}
\vspace{-2mm}
\begin{equation}
L_{\text{ent}} = - \frac{1}{|D^t|} \sum_{x_t\in D^t} \sum_{j=1}^{J} \delta_j(f^t(x_t)) \log \delta_j(f^t(x_t))
\end{equation}
\vspace{-0.12in}
\begin{equation}
L_{\text{div}} = \sum_{j=1}^{J} \hat{p}_j \log \hat{p}_j
\end{equation}
\vspace{-0.12in}
\begin{equation}
L_{pce} = -\beta \frac{1}{|D^t|}\sum_{x_t\in D^t} \sum_{j=1}^{J} 1_{[j = \hat{y}_t]}\log \delta_j(f^t(x_t))
\end{equation}
where $\hat{p_j} = \frac{1}{|D_t|} \sum_{x_t \in D_t} [\delta_j(f^{t}(x_t))]$ and $\beta >$ 0 is a balancing hyper-parameter. For the LoA baseline, each client k optimizes the $L_k^{LoA}$ objective independently leveraging the pre-trained server model as the source.

\subsection{Federated Learning}

\label{Pre:FL}
In FL, the goal is to solve the following optimization: 
\vspace{-2mm}
\begin{equation}
\underset{\mathbf{w}\in \mathbb{R}^n}{\arg\min} \ \left(l(\mathbf{w}) \triangleq \frac{1}{K} \sum_{k=1}^{K}l_{k}(\mathbf{w})\right)
\label{eq_f_def}
\vspace{-2mm}
\end{equation}

where $K$ is the number of clients, $ l_{k}(\mathbf{w}) $ is the client's task specific objective function and $\mathbf{w}$ denotes model parameters. It is solved in an iterative fashion spanning over $T$ communication rounds between the server and clients using methods such as FedAvg~\cite{mcmahan2017communication}.
\section{Method}

\begin{figure*}[h]
    \centering
        \includegraphics[scale=0.75]{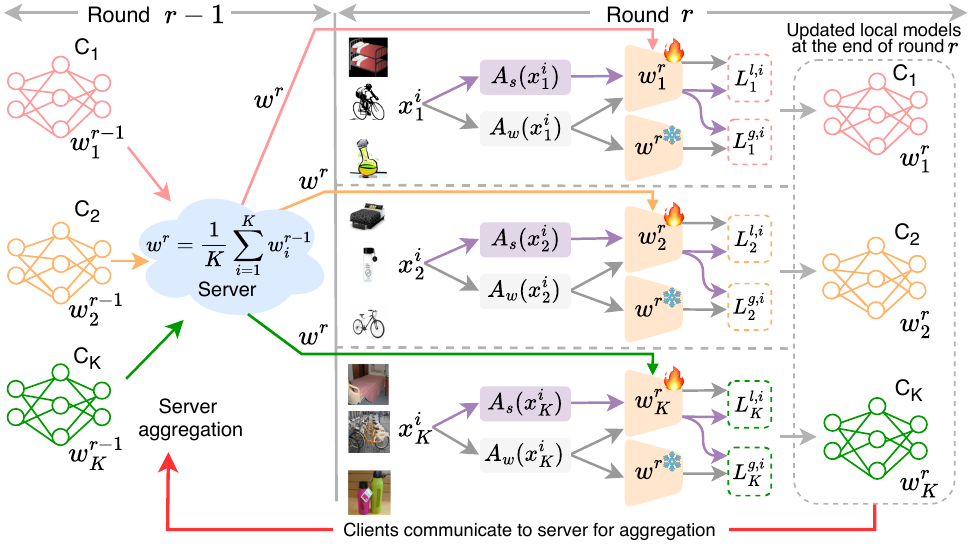}
        \caption{Overview of the proposed \textbf{FedSCAl framework with Server and Client Alignment (SCAl)} implementation. The server communicates the global model $\mathbf{w}^r$ to the clients, the clients then train their local models ($\mathbf{w}_k^r$) using a weak augmentation $A_w$ giving us the augmented image $A_w(x_k^i)$, and a strong augmentation $A_s$ from which we get the augmented image $A_s(x_k^i)$. 
        The clients compute the client alignment loss: $L_k^{l,i}$ and the server alignment loss: $L_k^{g,i}$ to perform the local training using the Eq.~\ref{client_loss}, and then communicate the updated local models $w_k^r$ to the server in the subsequent round for aggregation.}
        \label{intro_plot1}
\end{figure*}

\subsection{Problem Setup}
\label{sec:setup}
We assume a single server and $K$ clients and all the clients have the same model architecture. The server has access only to the pre-trained model trained on the source data with distribution $\mathbb{P}_{s}(x,y)$, and the $k^{th}$ client has access to the unlabelled dataset $\mathcal{D}_k$, 
drawn from the distribution 
$\mathbb{P}_{m}(x,y)$,
where $ m \in \{1,2,\dots, M \}$, and $M$ is the number of domains. Now, $\textstyle \mathbb{P}_{m}(x,y) = \mathbb{P}_{m}(y | x) \, \mathbb{P}_{m}(x)$ for features $x$ and labels $y$ belonging to domain $m$. 
Through this setup we tackle variations in the marginal distribution $\mathbb{P}_{m}(x)$ across the clients which we refer to as \textit{covariate shift} or domain gap i.e for any two clients $k$ and $l$, having sampled data from domains $k'$ and $l'$, we have $\mathbb{P}_{k'}(x) \neq \mathbb{P}_{l'}(x) $.
However, the problem is poorly defined if $\mathbb{P}_{s}(x,y)$ and $\mathbb{P}_{m}(x,y)$ are completely unrelated. So, we assume the source distribution $\mathbb{P}_{s}(x,y)$ and the domain distribution $\mathbb{P}_{m}(x,y)$ follow common domain adaptation assumptions~\cite{pmlr-v9-david10a,36364}. 
We consider the scenario where the datasets on any pair of clients belong to the same or different domains. 


\subsection{Pseudo-Label Estimation in Unsupervised Federated Settings}
\label{sec:Federated_pseudo_label}
FedSCAL aims to generate high-quality, robust pseudo labels with the help of the pre-trained server model assisting clients to train models on their unlabelled data via federation. To begin with, the estimation of pseudo-labels in a federated setting operates by deploying the LoA method to client $k$ which then yields the local client loss $L_{k}^{LoA}$. In each communication round $r$, the server broadcasts the global model $\mathbf{w}^r$ to participating clients who independently optimize their empirical local training loss $l_k(\mathbf{w})  \coloneqq \frac{1}{|\mathcal{D}_k|}\sum_{i=1}^{|\mathcal{D}_k|} L_k^{LoA}(\mathbf{w};x^i_k)$ where $x^i_k \in \mathcal{D}_k$.  
Each client $k$ optimizes the loss for a few local epochs $E$ and sends the model updates $\mathbf{w}_k^r$ to the server. In federated averaging, the server simply aggregates these model updates as $\mathbf{w}^{r+1} = \frac{1}{K}\sum_{k=1}^{K} \mathbf{w}_k^{r}$. 
This iterates until convergence and gives us the FedLoA baseline. However, the key bottleneck with FedLoA is the degradation in pseudo-label quality due to large domain gaps across the clients which results in client-drift. 
To address this we propose a novel Server and Client Alignment (SCAl) mechanism within the FedSCAl framework.   

\subsection{Improving Pseudo-label Accuracy via SCAl}
\label{sec:LGA}

We propose an alignment of the local client prediction with the prediction of the global model via our Server and Client Alignment (SCAl) mechanism. As aforementioned, clients with domain gap suffer from client drift, leading to overfitting on their local objective losses. 
To avoid this we focus on utilizing predictions consistently aligned with the global model. 
The alignment loss for client $k$ in a communication round is explained. Suppose $\delta(f({x}^i_k;w^r))$ denotes the prediction of the server model on a data point $x^i_k$ and similarly $\delta(f_{k}(x^i_k;w_k^r))$\footnote{For simplicity, the parameterizations \(w_r\) and \(w_r^k\) are omitted in subsequent discussions.} denotes the client model prediction on $x^i_k$. 
Here, $f_k(\cdot)$ represents the output of the classifier and $\delta(\cdot)$ is the softmax operator. We obtain the two predictions on single weakly augmented sample $A_w({x_{k}^i})$  of the original sample ${x^i_k}$ as $\delta(f_k(A_w({x_{k}^i})))$ and $\delta(f(A_w({x_{k}^i})))$ respectively. 
Similarly, we obtain the prediction of strongly augmented sample $A_s({x_k^i})$ as $\delta(f_k(A_s({x_k^i})))$.  We match the one-hot predictions of the weak augmentations from the client model $f_k$ (i.e, $\delta(f_k(A_w({x_{k}^i})))$ ) and the server model $f$ (i.e, $\delta(f(A_w({x_{k}^i})))$) using the alignment loss as described below. We denote the client alignment loss of the client model's weak and strong augmentations as  $L_k^{l,i}$, where
\begin{equation}
\begin{alignedat}{2}    
L_k^{l,i} &= \mathbb{I}_{\{\delta_{\text{max}}(f_k(A_w(x_k^i))) > \tau\}} \mathcal{D}_{KL}(p(x_k^i) || q(x_k^i)) \\
\end{alignedat}
\label{base_eq}
\end{equation}
and $\tau$ is the threshold,  $q(x_k^i) \coloneq \delta(f_k(A_s(x_k^i)))$ , $p(x_k^i) \coloneq \delta_{\text{oh}}(f_k(A_w(x_k^i)))$, and $\mathcal{D}_{KL}$ is the KL divergence. The threshold $\tau$ is employed to match the most confident predictions, this is discussed in the Sec.~\ref{adapt_thr}. 
$\delta_{oh}(f_k(A_w({x_{k}^i})))$ represents the one-hot prediction of the softmax vector $\delta(f_k(A_w({x_{k}^i})))$. $\delta_{max}$ denotes the maximum of the softmax outputs and $\mathbb{I}$ is the indicator function. 
Averaging across the batches yields the total client alignment loss as below.\footnote{If none of the predictions cross the threshold i.e., {$\sum_{i=1}^{B}{{\mathbb{I}_{\{\delta_{max}(f_k(A_w({x_{k}^i}))) > \tau\}}}} = 0$}, then $L^{l}_k$ is set to $0$}
\begin{equation}
L^{l}_k = \frac {1}{\sum_{i=1}^{B}{{\mathbb{I}_{\{\delta_{max}(f_k(A_w({x_{k}^i}))) > \tau\}}}}} \sum_{i=1}^{B}{L_k^{l,i}}
\label{loc:lga}
\end{equation}

where $B$ denotes the size of the mini-batch.


\noindent Similarly, the server alignment loss denoted as $L_k^{g,i} $ is 

\vspace{-3mm}
\begin{equation}
\begin{alignedat}{2}
L_k^{g,i} &= \mathbb{I}_{\{\delta_{\text{max}}(f(A_w(x_k^i))) > \tau\}} \, \mathcal{D}_{KL}(r(x_k^i) || q(x_k^i))
\end{alignedat}
\end{equation}
where $r(x_k^i) \coloneq \delta_{\text{oh}}(f(A_w(x_k^i)))$. 
Averaging across the batches yields the total server alignment loss\footnote{ If none of the samples cross the threshold i.e., ${\sum_{i=1}^B{{\mathbb{I}_{\{\delta_{max}(f(A_w({x_{k}^i}))) > \tau\}}}}} = 0$, then $L^{g}_k$ is set to $0$.} as
\begin{equation}
L^{g}_k = \frac {1}{\sum_{i=1}^B{{\mathbb{I}_{\{\delta_{max}(f(A_w({x_{k}^i}))) > \tau\}}}}} \sum_{i=1}^{B}L_k^{g,i}
\label{glob:lga}
\end{equation}
The overall alignment loss is the sum of the server and client alignment losses and is given by 

\begin{equation}
 L_k^{SCAl} = \lambda_{l}*L_k^{l} + \lambda_{g}*L_k^{g}
 \label{const_loss}
\end{equation}

\begin{figure}[htbp]
  \centering  
  \begin{subfigure}[t]{0.36\textwidth}    
    \centering  
  \includegraphics[width=\linewidth]{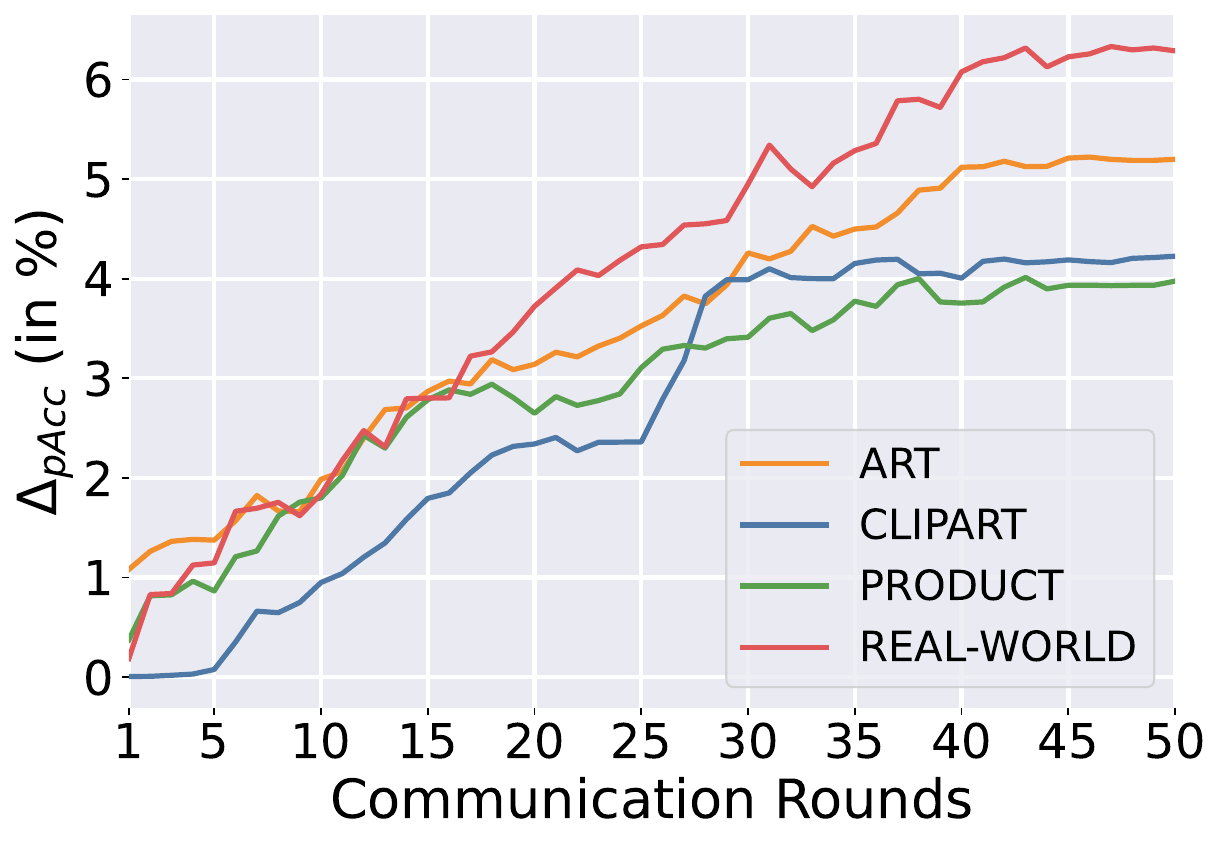}
    \caption{Pseudo-label prediction accuracy difference $\Delta_{pAcc}$ (Eq.~\ref{pacc_eq}) improves with the proposed SCAl mechanism  on the Office-Home dataset, when the training is initialized with the server model pre-trained on different source domains (Art, ClipArt, Product and Real-World).}
    \label{pseudo_acc}
  \end{subfigure}
  \begin{subfigure}[t]{0.4\textwidth}
    \centering 
    \includegraphics[width=\linewidth]{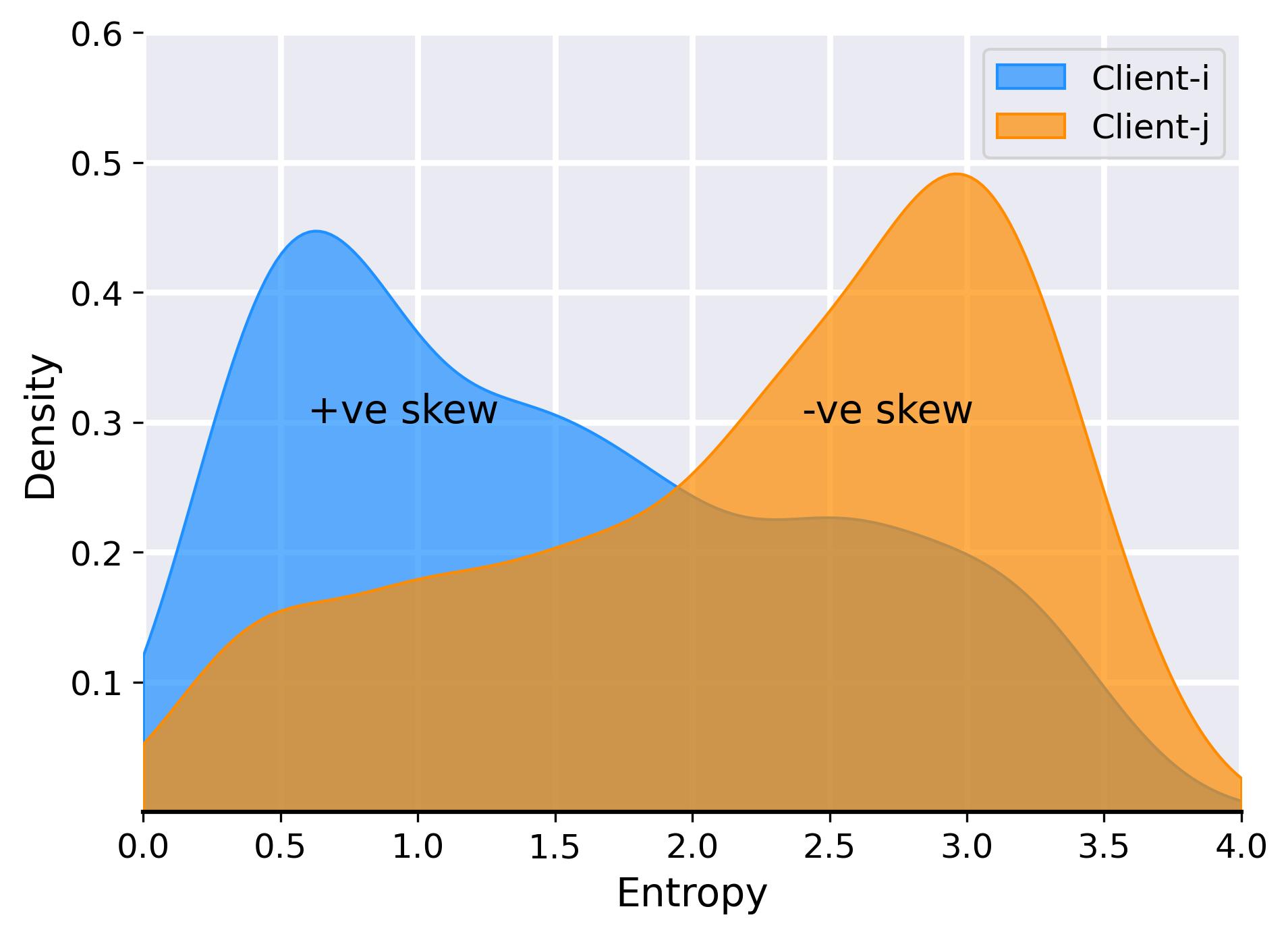}
    \caption{
    Entropy density of the clients predictions on Office-Home dataset, calculated via (Eq.~\ref{entropy_eq}), exhibiting skewness ($\gamma$); with negative $\gamma$ (e.g., $\gamma=-0.1$) indicating higher entropy and lower prediction confidence, while positive $\gamma$ (e.g., $\gamma=0.15$) reflecting lower entropy and higher prediction confidence.
    }
    \label{fig:skewness_demo_plot}
  \end{subfigure}
  \caption{Pseudo-label Accuracy difference and visualization of Entropy Density of Clients Predictions.}
  \label{fig_method2}
 \vspace{-0.1in}  
\end{figure}

Finally, our local client loss takes the form, here $L_k^{LoA}$ refers to client $k$ adapting the Eq.~\ref{eq_shot}.  
\begin{equation}
l_k(\mathbf{w}) = L_k^{LoA} + L_k^{SCAl}
\label{client_loss}
\end{equation}
We demonstrate the impact of the client and server alignment loss. Figure~\ref{pseudo_acc} shows the pseudo-label accuracy difference $\Delta_{pAcc}$ (Eq.~\ref{pacc_eq}) with and without SCAl loss.
\begin{equation}
\Delta_{pAcc} = pAcc(\text{FedSCAl}) - pAcc(\text{FedLoA})
\label{pacc_eq}
\end{equation}

where $pAcc (\text{FedSCAl})$ gives the pseudo-label accuracy 
when we use our proposed FedSCAl with Server and Client Alignment (SCAl) mechanism
and  $pAcc(\text{FedLoA})$ gives the pseudo-label accuracy for the adaptation of LoA method to a federated setup. In Figure~\ref{pseudo_acc}, it can be seen that when we deploy our FedSCAl framework with the SCAl loss (Eq.~\ref{const_loss}), the pseudo-label accuracy 
keeps improving over the communication rounds. 

\renewcommand{\arraystretch}{1.38}
\begin{table*}[!t] 
\centering
\caption{Results on \textbf{Office-Home} (Art: \textbf{A}, ClipArt: \textbf{C}, Real-World: \textbf{R}, and Product: \textbf{P}) Dataset, while the initial server model is pre-trained on one of the $4$ domains and the clients are distributed with the remaining $3$ domains such that each client contains the subset of exactly one domain. We report the \textbf{Accuracy(\%)} for each target domain and the average of all the domains for each pre-trained model. It can be seen that FedSCAl improves the performance significantly 
across all the domains. Here `$\rightarrow$' denotes Source $\rightarrow$ Target.}
\vspace*{-1mm}
\scalebox{0.7}{
\begin{tabular}{l|ccc|c|ccc|c|ccc|c|ccc|c}
\toprule
\multirow{2}{*}{Method} & \multicolumn{4}{c|}{\makecell{(a) Initial Sever Model is \\pre-trained on \textbf{\texttt{Art}}}} & \multicolumn{4}{c|}{\makecell{(b) Initial Sever Model is \\pre-trained on \textbf{\texttt{Clipart}}}} & \multicolumn{4}{c|}{\makecell{(c) Initial Sever Model is \\pre-trained on \textbf{\texttt{Product}}}} & \multicolumn{4}{c}{\makecell{(d) Initial Sever Model is \\pre-trained on \textbf{\texttt{Real-World}}}}\\

& \textbf{A$\rightarrow$C} & \textbf{A$\rightarrow$P} & \textbf{A$\rightarrow$R} & \textbf{Avg.} & \textbf{C$\rightarrow$A} & \textbf{C$\rightarrow$P} & \textbf{C$\rightarrow$R} & \textbf{Avg.} & \textbf{P$\rightarrow$A} & \textbf{P$\rightarrow$C} & \textbf{P$\rightarrow$R} & \textbf{Avg.} & \textbf{R$\rightarrow$A} & \textbf{R$\rightarrow$C} & \textbf{R$\rightarrow$P} & \textbf{Avg.}\\ \midrule
 
LoA & 44.95 & 79.75 & 84.86 & 69.85 & 73.86 & 80.45 & 84.03 & 79.45 & 69.54 & 46.42 & 85.98 & 67.31 & 74.85 & 48.12 & 85.03 & 69.34\\ \midrule
FedLoA & 56.69 & 83.56 & 86.38 & 75.55 & 78.97 & 84.33 & 86.24 & 83.18 & 75.92 & 52.82 & 87.08 & 71.94 & 77.55 & 55.01 & 86.68 & 73.08\\ \midrule
FedProx &57.95	&83.48&	86.26	&75.90	&79.81	&85.10&	86.12	&83.67 &73.98	&54.93	&85.71	&71.54	&77.12	&57.06	&85.92	&73.38\\ \midrule
FedMOON & 56.23	&85.08	&87.13	&76.15	&79.42	&85.67	&87.34	&84.15 &75.66	&53.34	&86.78	&71.93	&78.86	&55.26	&86.59	&73.57\\ \midrule
FedWCA &53	&84.02&	86.72&	74.58&	78.88&	84.83	&86.68&	83.46&	75.43&	52.39	&86.95&	71.59&	76.59&	54.25&	86.12&	72.32\\ \midrule
LADD & 37.50 & 65.21 & 71.23 & 57.98 & 53.90 & 66.24 & 70.42 & 63.52 & 61.21 & 42.46 & 79.09 & 60.92 & 66.30 & 40.80 & 78.36 & 61.82\\ \midrule 
Dual Adapt & 45.85 & 76.56 & 84.11 & 68.84 & 70.58 & 78.06 & 81.99 & 76.88 & 68.92 & 44.33 & 84.73 & 65.99 & 73.79 & 46.17 & 84.1 & 68.02\\ \midrule 
\cellcolor{cyan!8} {\textbf{FedSCAl} (Ours)} & \cellcolor{cyan!8} {{61.22}} & \cellcolor{cyan!8} {{87.16}} & \cellcolor{cyan!8} {{88.76}} & \cellcolor{cyan!8}{\textbf{79.05}} & \cellcolor{cyan!8}{{81.13}} & \cellcolor{cyan!8}{{88.19}} & \cellcolor{cyan!8}{{88.69}} & \cellcolor{cyan!8}{\textbf{86.01}} & \cellcolor{cyan!8}{{79.39}} & \cellcolor{cyan!8}{{58.48}} & \cellcolor{cyan!8}{{88.81}} & \cellcolor{cyan!8}{\textbf{75.56}} & \cellcolor{cyan!8}{{79.68}} & \cellcolor{cyan!8}{{60.96}} & \cellcolor{cyan!8}{{89.12}} & \cellcolor{cyan!8}{\textbf{76.59}}\\ \bottomrule

\end{tabular}
}
\label{tab1:office_home}
\end{table*}

\subsection{Adaptive Thresholding}
 \label{adapt_thr}
 In Eq.~\ref{base_eq}, we employ the threshold $\tau$ to match the most confident predictions. However, setting a fixed value for $\tau$ during the federated training may encounter certain drawbacks. Firstly, during the initial phases of training, the client models may not be very confident in their predictions, necessitating the initial relaxation of the threshold, followed by tightening as the client models' predictions grow in confidence. Another issue may arise due to client data heterogeneity; client models deployed on certain domains may have more or less confidence in their predictions than client models deployed on other domains. This motivates us to employ adaptive thresholding during the training cycles.

We exploit the skewness of the entropy (Eq.~\ref{entropy_eq}) distribution of the predictions on the client data samples, to dynamically adjust the threshold.
\begin{equation}
\small
H(\delta(f_k(x^i_k))) = - \sum_{j} \delta_j(f_k(x^i_k))   \log \left( \delta_j(f_k(x^i_k)) \right)
\label{entropy_eq}
\vspace{-2mm}
\end{equation}

where $\delta_j$ is the softmax confidence that  $x^i_k$ belongs to class $j$.
We denote the $\mu_H$, $M_H$, and $\sigma_H$ as the mean, median, and standard deviation of the entropy distribution, respectively.
The skewness $\gamma$ (~\cite{Pearson1894-dv} and~\cite{De_la_Rubia2023-wb}) is given by following
\begin{equation}
\gamma = 3*\frac{\mu_H - M_H}{\sigma_H}
\label{skewness_eq}
\end{equation}
A positive value for the skewness $\gamma$ (Figure.~\ref{fig:skewness_demo_plot}) implies an entropy distribution skewed towards lower entropy values. This signals that the client model is confident in its predictions. Likewise, a negative value for $\gamma$ would imply that the distribution is skewed towards higher entropy values meaning the client model is less confident in its predictions. For further enhancing the pseudo-label estimation using FedSCAl, we tighten the threshold for confident client models and relax it for the less confident ones. We start with an initial threshold of $\tau_{init}$ and then dynamically update $\tau$ during training as
\vspace{-0.1in}
\begin{equation}
 \tau = \tau_{init} + max(\gamma_L,min(\gamma,\gamma_H))  \end{equation} 
 To ensure that we are not left with too few prediction class labels for a very high threshold $\tau$ while also ensuring we do not sample noisy predictions for a very small $\tau$, we clip the threshold in the range of [$\gamma_{L}$, $\gamma_H$]. The detailed algorithm is provided in Sec. 2 of the Supplementary. The impact of the adaptive threshold is demonstrated in Table~\ref{table:threshold}.

\vspace{-0.05in}
\section{Experiments}
\label{sec:Exp}
\subsection{Datasets and Setup}
For all our experiments, we choose an ImageNet pre-trained ViT small model. We consider the
cross-silo setting inspired from FedBN~\cite{li2021fedbn} and pFedPG~\cite{yang2023efficientmodelpersonalizationfederated}, where a data domain is assigned to a client. We consider an even difficult setting, wherein each domain is distributed across a small number of clients and set a participation rate ($<$\:1) allowing only a few clients to participate during each communication round. We consider  Office-Home and DomainNet datasets for our experiments.
\textbf{Office-Home}~\cite{venkateswara2017deep} dataset has four domains namely \texttt{Art}, \texttt{Clipart}, \texttt{Product}, and \texttt{Real-World}.
While we consider one domain as a source, the other $3$ domains are distributed among $15$ clients ($5$ each for a particular domain). The participation rate is set to 0.3 for the experiment on \textit{Office-Home} dataset. 
\textbf{Domain-Net}~\cite{Gong2012GeodesicFK} dataset has six domains namely \texttt{Clipart}, \texttt{Infograph}, \texttt{Painting}, \texttt{Quickdraw}, \texttt{Real}, and \texttt{Sketch}. We follow the protocol of~\cite{li2021fedbn} for adapting the Domain-Net dataset to the federated setup.
While we consider one domain as a source, the other $5$ domains are distributed among $10$ clients ($2$ each for a particular domain). The participation rate is set to $0.3$. Local training epochs for each client are kept at 5 for both datasets with a learning rate of $0.3$. 

\textbf{\textit{Augmentations}:} Our weak augmentations consist of resizing the image to a uniform size, followed by a random resized crop and horizontal flip, which are then transformed to tensors and normalized. For strong augmentations, we randomly choose a pair of augmentations to be applied sequentially to the image from a diverse pool. This augmentation pool includes transformations as used in FixMatch~\cite{sohn2020fixmatch}. 
\vspace{-0.05in}
\subsection{Baselines and Evaluation}  

To establish baselines, we include LoA (Section~\ref{Pre:sfda}) and FedLoA (Section~\ref{sec:Federated_pseudo_label}), and compare against LADD~\cite{shenaj2023learning}, DualAdapt~\cite{9706703}, FedProx~\cite{li2020federated}, FedMOON~\cite{li2021model}, FedWCA~\cite{mori2025federated}, and our proposed FedSCAl. We report the best average accuracy (\%), by averaging the accuracies of clients belonging to each domain. By testing on the client data, our evaluation captures both performance and generalization capabilities. The hyper-parameter settings and implementation details are provided in the Sec. 7 of the supplementary material. 



\begin{table*}[t]
\renewcommand{\arraystretch}{1.45}
\caption{Results on \textbf{DomainNet} Dataset (Clipart: \textbf{C}, Infograph: \textbf{I}, Product: \textbf{P}, Sketch: \textbf{S}, Quickdraw: \textbf{Q}, and Real: \textbf{R}), while the initial server model is pre-trained on one of the $6$ domains and clients are distributed with the remaining $5$ domains such that each client contains the subset of exactly one domain.
We report \textbf{accuracy(\%)} for each target domain and the average of all domains for each pre-trained model. 
It is evident that adding our SCAl regularizer boosts the performance 
across all the domains. Here `$\rightarrow$' denotes Source $\rightarrow$ Target.}
\vspace*{-1mm}
\centering
\scalebox{0.63}{
\begin{tabular}{l|ccccc|c|ccccc|c|ccccc|c}
\toprule
\multirow{2}{*}{Method} & \multicolumn{6}{c|}{\makecell{(a) Initial Sever Model is pre-trained on \textbf{\texttt{Clipart}}}} & \multicolumn{6}{c|}{\makecell{(b) Initial Sever Model is pre-trained on \textbf{\texttt{Infograph}}}} & \multicolumn{6}{c}{\makecell{(c) Initial Sever Model is pre-trained on \textbf{\texttt{Painting}}}}\\

& \textbf{C$\rightarrow$I} & \textbf{C$\rightarrow$P} & \textbf{C$\rightarrow$Q} & \textbf{C$\rightarrow$R} & \textbf{C$\rightarrow$S} & \textbf{Avg.} & \textbf{I$\rightarrow$C} & \textbf{I$\rightarrow$P} & \textbf{I$\rightarrow$Q} & \textbf{I$\rightarrow$R} & \textbf{I$\rightarrow$S} & \textbf{Avg.} &
\textbf{P$\rightarrow$C} & \textbf{P$\rightarrow$I} & \textbf{P$\rightarrow$Q} & \textbf{P$\rightarrow$R} & \textbf{P$\rightarrow$S} & \textbf{Avg.} \\ \midrule
LoA & 49.00 & 93.90 & 70.5 & 95.57 & 86.77 & 79.15 & 80.24 & 92.21 & 54.32 & 94.35 & 82.30 & 80.80&79.93 & 49.43 & 33.22 & 94.88 & 83.30 & 68.15 \\ \midrule
FedLoA & 52.56 & 93.69 & 73.49 & 96.13 & 87.03 & 80.58 & 80.66 & 91.62 & 70.79 & 95.95 & 83.86 & 84.58 & 81.93 & 54.42 & 68.91 & 95.80 & 81.06 & 76.42 \\ \midrule
FedProx &56.78	&92.75	&70.35	&95.89	&87.95	&80.74	&80.63	&91.58	&61.71	&95.86	&83.59	&82.67 & 79.15&	54.50&	52.44	&95.17	&82.54&	72.81\\ \midrule

FedMOON &56.31	&93.06	&64.75&	96.21&	88.15&	79.7&	81.77&	92.46&	56.25&	96.34&	83.00&	81.97 & 78.04	&52.92&	36.24	&96.09	&82.58&	69.17\\ \midrule
FedWCA &57.13&	93.77	&73.37&	96.19	&88.5	&81.79&	80.48	&92.20	&67.44&	96.21	&82.48	&83.76	&80.36	&54.49	&64.03	&95.94&	80.80&	75.12\\ \midrule
LADD & 53.66 & 91.28 & 34.69 & 95.83 & 82.72 & 71.64 & 66.11 & 80.52 & 26.77 & 85.21 & 65.59 & 64.84 & 60.94 & 36.67 & 10.14 & 94.33 & 72.86 & 54.99\\ \midrule
DualAdapt & 54.46 & 90 & 35.36 & 95.59 & 85.33 & 72.15 & 67.35 & 81.09 & 19.63 & 86.52 & 69.37 & 64.79 & 73.05 & 47.17 & 10 & 95.01 & 80.29 & 61.1\\ \midrule

\cellcolor{cyan!8}{\textbf{FedSCAl} (Ours)} & \cellcolor{cyan!8}{{60.1}} & \cellcolor{cyan!8}{{94.78}} & \cellcolor{cyan!8}{{78.8}} & \cellcolor{cyan!8}{{96.53}} & \cellcolor{cyan!8}{{92.72}} & \cellcolor{cyan!8}{\textbf{84.58}} & \cellcolor{cyan!8}{{87.09}} & \cellcolor{cyan!8}{{94.78}} & \cellcolor{cyan!8}{{76.01}} & \cellcolor{cyan!8}{{96.42}} & \cellcolor{cyan!8}{{88.94}} & \cellcolor{cyan!8}{\textbf{88.65}} & \cellcolor{cyan!8}{{85.96}} & \cellcolor{cyan!8}{{59.23}} & \cellcolor{cyan!8}{{61.43}} & \cellcolor{cyan!8}{{96.60}} & \cellcolor{cyan!8}{{88.78}} & \cellcolor{cyan!8}{\textbf{78.4}}\\ \bottomrule

\multirow{2}{*}{Method} & \multicolumn{6}{c|}{(d) Initial Sever Model is pre-trained on \textbf{\texttt{Quickdraw}}} & \multicolumn{6}{c|}{(e) Initial Sever Model is pre-trained on \textbf{\texttt{Real}}} & \multicolumn{6}{c}{(f) Initial Sever Model is pre-trained on \textbf{\texttt{Sketch}}}\\

& \textbf{Q$\rightarrow$C} & \textbf{Q$\rightarrow$I} & \textbf{Q$\rightarrow$P} & \textbf{Q$\rightarrow$R} & \textbf{Q$\rightarrow$S} & \textbf{Avg.} & \textbf{R$\rightarrow$C} & \textbf{R$\rightarrow$P} & \textbf{R$\rightarrow$Q} & \textbf{R$\rightarrow$I} & \textbf{R$\rightarrow$S} & \textbf{Avg.} & \textbf{S$\rightarrow$C} & \textbf{S$\rightarrow$P} & \textbf{S$\rightarrow$Q} & \textbf{S$\rightarrow$R} & \textbf{S$\rightarrow$I} & \textbf{Avg.} \\ \midrule
LoA & {69.68} & 14.91 & 39.15 & 56.38 & {75.25} & 51.07 & 70.46 & 85.87 & 59.43 & 46.51 & 73.90 & 67.23 & 85.20 & 92.70 & 77.60 & 95.47 & 48.58 & 79.91\\ \midrule
FedLoA & 45.21 & 29.56 & 50.02 & 64.08 & 52.35 & 48.24 &  72.58 & 86.09 & 70.43 & 45.57 & 76.09 & 70.15 & 86.33 & 93.38 & 72.91 & 96.29 & 56.02 & 80.99\\ \midrule
FedProx & 59.74&	35.76 &	51.33	&57.34	&60.86&	53.01 & 71.79	&86.75	&51.97	&48.55&	73.49	&66.51&	87.2&	91.62&	67.72&	95.57	&53.94&	79.21\\ \midrule
FedMOON & 60.34& 30.44	&	52.90&	66.35&	56.41&	53.29 & 76.73&	90.24	&58.71&	53.35&	79.6&	71.73&	86.48&	93.64&	68.83&	96.18&	54.75&	79.98\\ \midrule
FedWCA &41.49	&32.70&	46.22&	54.74&	43.29&	43.68&	72.51&	87.63&	64.51&	48.63&	74.83&	69.62&	86.90&	92.80&73.04	&95.99	&51.33&	80.02\\ \midrule
LADD &  52.72 & 25.94 & 37.19 & 50.80 & 52.95 & 43.92 & 51.79 & 74.70 & 10.90 & 32.01 & 61.64 & 46.21 & 79.54 & 89.83 & 42.67 & 95.26 & 53.49 & 72.15 \\ \midrule   
DualAdapt & 59.78 & 22.02 & 40.68 & 51.85 & 49.58 & 44.78 & 69.72 & 86.14 & 13.43 & 46.54 & 73.36 & 57.84 & 86.29 & 91.67 & 39.53 & 95.76 & 56.08 & 73.86\\ \midrule
\cellcolor{cyan!8}{\textbf{FedSCAl} (Ours)} & \cellcolor{cyan!8}{{65.37}} & \cellcolor{cyan!8}{{40.15}} & \cellcolor{cyan!8}{{65.74}} & \cellcolor{cyan!8}{{66.03}} & \cellcolor{cyan!8}{{65.75}} & \cellcolor{cyan!8}{\textbf{60.61}} & \cellcolor{cyan!8}{{81.74}} & \cellcolor{cyan!8}{{92.33}} & \cellcolor{cyan!8}{{76.84}} & \cellcolor{cyan!8}{{53.45}}  & \cellcolor{cyan!8}{{86.03}} & \cellcolor{cyan!8}{\textbf{78.08}} & \cellcolor{cyan!8}{{89.96}} & \cellcolor{cyan!8}{{95.02}} & \cellcolor{cyan!8}{{77.61}} & \cellcolor{cyan!8}{{96.53}} & \cellcolor{cyan!8}{{58.89}} & \cellcolor{cyan!8}{\textbf{83.60}} \\ \bottomrule
\end{tabular}
}
\label{tab:domain_net}
\vspace{-4mm}
\end{table*}
\subsection{Results and Discussion}
\textbf{\textit{Office-Home}:} In Table~\ref{tab1:office_home}, we present the results for the Office-Home dataset, for all the combinations of server and client domains. For example, Table~\ref{tab1:office_home}(a) shows that the server has the model pre-trained on the Art domain, and the clients are distributed with Clipart, Product, and Real-World data. 
 It can be seen that  LADD performs inferior to other baseline methods. The LoA baseline also performs inferior to baselines such as FedLoA, key issue being the training on data limited to a single client. In the case of FedLoA, the performance degrades due to client-drift. FedSCAl performs the best as it mitigates the client-drift by employing the Server and Client Alignment(SCAl) mechanism and also improves the generalization performance.
FedSCAl achieves $3.50\%$ improvement over the FedLoA baseline for the Art pre-trained server model (Table~\ref{tab1:office_home}(a)), and $3.51\%$ improvement for the Real-World pre-trained server model (Table~\ref{tab1:office_home}(d)). For the Clipart and Product pre-trained server models, FedSCAl yields average performance gains of $2.83\%$ and $3.62\%$, respectively, compared to FedLoA (Table~\ref{tab1:office_home}(b,c)). Table~\ref{tab1:office_home} further compares FedSCAl with FedProx~\cite{li2020federated} and Model Contrastive (MOON)~\cite{li2021model}. FedSCAl consistently outperforms both FedProx and FedMOON across all target domains. Specifically, with the Art pre-trained server model, FedSCAl delivers an average improvement of $3.15\%$ over FedProx and $2.90\%$ over FedMOON. Comparable improvements are observed over other methods, including FedWCA, LADD, and Dual Adapt.


\textbf{\textit{DomainNet}:} The results on the DomainNet dataset are presented in Table~\ref{tab:domain_net}. With the Clipart pre-trained server model (Table~\ref{tab:domain_net}(a)), the proposed method achieves an average improvement of $4\%$ over the FedLoA baseline, with consistent gains across all target domains and a maximum increase of $7.54\%$ on the Infograph domain. For the Infograph pre-trained server model (Table~\ref{tab:domain_net}(b)), our approach yields an average improvement of $4.07\%$ over FedLoA, reaching up to $6.43\%$ on the Clipart domain. Using the Painting pre-trained server model (Table~\ref{tab:domain_net}(c)), the method delivers an average improvement of $1.98\%$, peaking at $7.72\%$ on the Sketch domain. When initialized with the Quickdraw pre-trained server model (Table~\ref{tab:domain_net}(d)), the proposed method achieves a substantial average improvement of $9.54\%$ over the LoA baseline, showing its effectiveness. Comparable performance gains are observed with server models pre-trained on other domains such as Real and Sketch, demonstrating the robustness and adaptability of the approach across diverse visual domains. Furthermore, FedSCAl consistently outperforms competing federated learning algorithms, including FedProx and FedMOON, in all evaluated settings. For example, with the Real domain pre-trained server model, FedSCAl achieves an average improvement of $11.57\%$ over FedProx and $6.35\%$ over FedMOON. These consistent improvements across initializations, including Painting, Quickdraw, and Clipart, show the generalization ability of the proposed method for federated learning with heterogeneous data under the FFreeDA setup. Office-31 results are in Sec. 6 of the supplementary.

\subsection{Analysis of SCAl Mechanism}
\textbf{\textit{Server and Client Alignment Losses}:} We conduct a detailed ablation study on our proposed SCAl loss. We consider the Office-Home dataset for this experiment. As illustrated in Table~\ref{tab1:ablation}, each component of our method contributes significantly to the performance gain. Introducing the server alignment loss boosts the FedSCAl performance to around $\approx1.90\%$ over FedSCAl with only the client alignment loss.
\begin{table}[htp]
\caption{Ablation study of SCAl Losses on \textbf{Office-Home} dataset.}
\centering
\vspace{-3pt}
\resizebox{0.45\textwidth}{!}{%
\begin{threeparttable}
\begin{tabular}{c|c|c|c|c|c|c}
\toprule
 \textbf{$L_k^{l}$} & \textbf{$L_{k}^{g}$} & A$\rightarrow$\{C,\:P,\:R\} & C$\rightarrow$\{A,\:P,\:R\} & P$\rightarrow$\{C,\:A,\:R\} & R$\rightarrow$\{C,\:P,\:A\} & { \# \textbf{Avg.}}\\ \midrule

 \crossmark[red, scale=1] & \crossmark[red, scale=1] & 75.55 & 83.18 & 71.94 & 73.08 & 75.93\\ 
 \greencheck & \crossmark[red, scale=1] & 77.13 & 84.32 & 74.12 & 74.25 & 77.46\\ 
 \crossmark[red, scale=1] & \greencheck & 77.33 & 84.63 & 74.28 & 74.41 & 77.66\\ 
 \cellcolor{cyan!8}\greencheck & \cellcolor{cyan!8}\greencheck & \cellcolor{cyan!8}{\textbf{79.05}} & \cellcolor{cyan!8}{\textbf{86.01}} & \cellcolor{cyan!8}{\textbf{75.56}} & \cellcolor{cyan!8}{\textbf{76.59}} & \cellcolor{cyan!8}{\textbf{79.30}}\\ \bottomrule

\end{tabular}
\begin{tablenotes}
    \centering
    \item ‘$\rightarrow$’ denotes Source $\rightarrow$ Target; Art (A), ClipArt (C), Product (P), and Real-World (R).
    \end{tablenotes}
\end{threeparttable}}
\vspace{-10pt}
\label{tab1:ablation}
\end{table}

\begin{figure}[htp]
\centering
\includegraphics[scale=0.33]{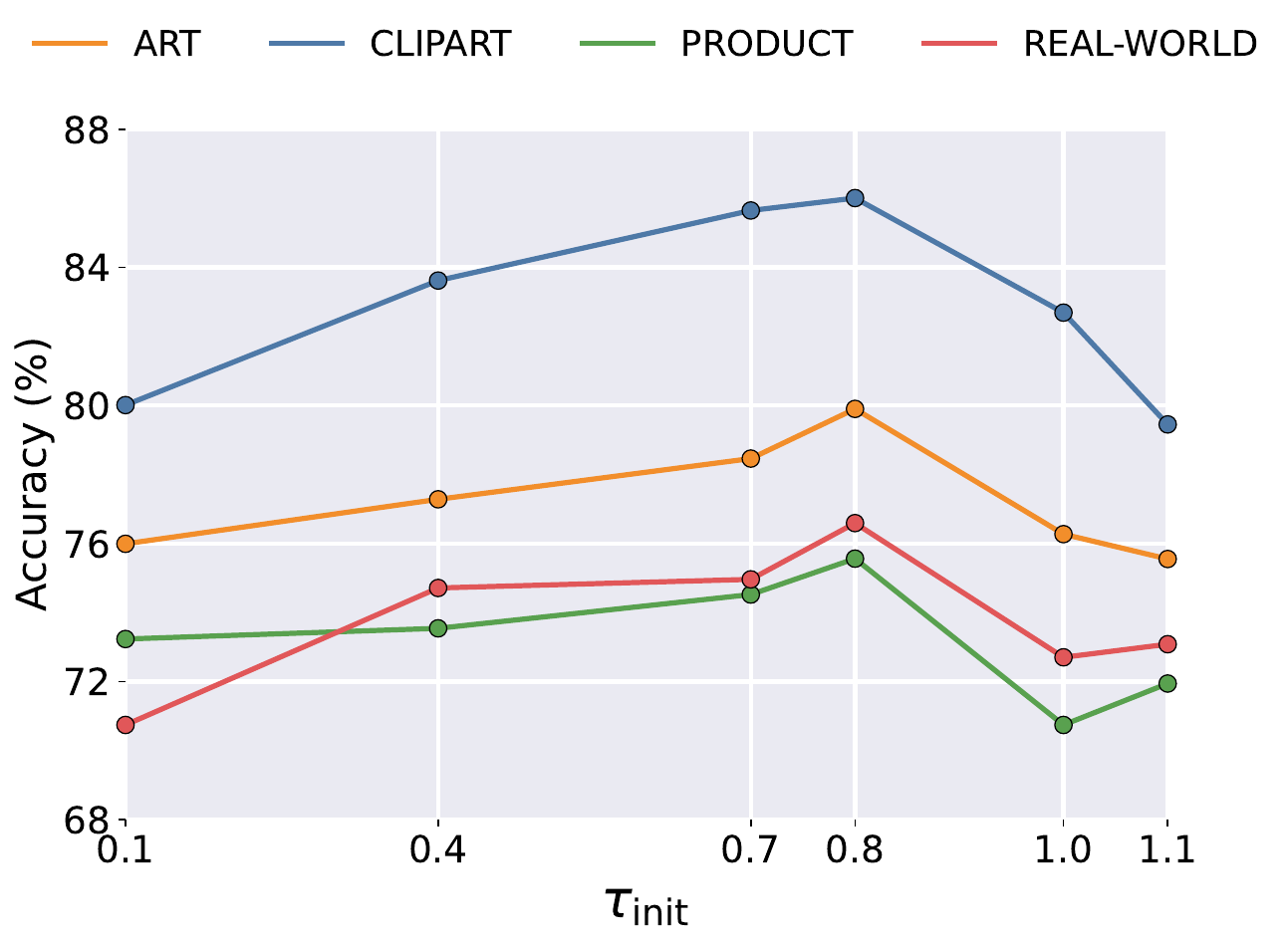}
\vspace{-2mm}
    \caption{Effect of varying $\tau_{init}$ on the average accuracy attained by all the clients when the training is initialized with the server model pre-trained on Art, Clipart, Product, and Real-World.}
    \label{init_thr_plot}
    \vspace{-3mm}
\end{figure}

\textbf{\textit{Varying the threshold $\tau_{init}$}:} For all experiments, we set $\tau_{init}=0.8$, $\gamma_L = -0.1$ and $\gamma_H = 0.15$ as a design choice. This configuration enables fine-grained threshold adjustments, resulting in lower and upper bounds of $0.7$ and $0.95$, respectively. If larger clipping values were used (e.g., $\gamma_L = \gamma_H = 0.6$), the threshold would span from $0.2$ to $1.4$. At the upper bound of $1.4$, the SCAl loss remains inactive, whereas at the lower bound of $0.2$, it may be triggered by noisy predictions, potentially degrading performance. The rationale for selecting $\tau_{\mathrm{init}} = 0.8$ was to mitigate the influence of noisy samples at the onset of training. To illustrate the impact of varying $\tau_{\mathrm{init}}$ while keeping $\gamma_L$ and $\gamma_H$ fixed, we present an experiment in Fig.~\ref{init_thr_plot} using the Office-Home dataset. In this figure, we plot the average accuracy of all the clients when the training is initialized with the server model pre-trained on different source domains such as  Art, Clipart, Product, and Real-World. 
\vspace{-4mm}

\textbf{\textit{Adaptive Vs Fixed Threshold}:} 
In Table~\ref{table:threshold}, we report average accuracies on \textit{Domain-Net} for the proposed adaptive threshold compared to the fixed threshold. We demonstrate an improvement of 1.81\% when the initial server model is pre-trained with Clipart and an improvement of 1.98\% when the initial server model is pre-trained using Quickdraw with adaptive thresholding. Similar improvements can be observed for other domains with adaptive thresholding improving performance for all the transfer settings.
\begin{table}[t]
\centering
\caption{SCAl with Adaptive \emph{vs} Fixed Thresholding. 
}
\vspace{-1mm}
\resizebox{0.47\textwidth}{!}{%
\begin{threeparttable}
\begin{tabular}{c|c|c|c|c|c|c}
\toprule
\multirow{2}{*}{Thresholding} & C$\rightarrow$ & I$\rightarrow$ & P$\rightarrow$\ & Q$\rightarrow$ & R$\rightarrow$ & S$\rightarrow$ \\
& \{R,I,P,Q,S\} & \{R,C,P,Q,S\} & \{R,I,C,Q,S\} & \{R,I,P,C,S\} & \{C,I,P,Q,S\} & \{R,I,P,C,Q\}\\ \midrule
Fixed & 82.77 & 87.77 & 78.19 & 58.63 & 75.81 & 83.4 \\ 
\cellcolor{cyan!8}{\textbf{Adaptive}} & \cellcolor{cyan!8}{\textbf{84.58}} & \cellcolor{cyan!8}{\textbf{88.65}} & \cellcolor{cyan!8}{\textbf{78.40}} & \cellcolor{cyan!8}{\textbf{60.61}} & \cellcolor{cyan!8}{\textbf{78.08}} & \cellcolor{cyan!8}{\textbf{83.60}} \\ \bottomrule
\end{tabular}

\begin{tablenotes}
    \centering
    \item ‘$\rightarrow$’ denotes Source $\rightarrow$ Target; Clipart: \textbf{C}, Infograph: \textbf{I}, Product: \textbf{P}, Sketch: \textbf{S}, Quickdraw: \textbf{Q}, and Real: \textbf{R} 
\end{tablenotes}
\end{threeparttable}}
\vspace{-2mm}
\label{table:threshold}
\end{table}
\textbf{\textit{On the Choice of LoA}:} 
In our main experiments, we used LoA method as SHOT due to its simplicity and as well as to ensure fair comparison with SOTA methods such as FedWCA~\cite{mori2025federated}, which are designed atop of it.
Here, we showcase experiments with alternative LoA methods such as BMD~\cite{qu2022bmd} and UCon~\cite{xu2025revisiting} in Tables~\ref{tab:bmd_loa} and~\ref{tab:ucon_loa}, respectively. This demonstrates that FedSCAl framework works independently of the underlying LoA method. Due to space constraints, we provide only the average accuracy from each source; the detailed splits of every source-target pairs and the Domain-Net results are provided in the Sec. 4 of the supplementary material.
\vspace{-4pt}
\begin{table}[htp]
\caption{Comparison of FedSCAl and FedLoA using \textbf{BMD} as the underlying LoA method.}
\centering
\vspace{-3pt}
\resizebox{0.47\textwidth}{!}{%
\begin{threeparttable}
\begin{tabular}{l|c|c|c|c|c}
\toprule
Method & A$\rightarrow$\{C,\:P,\:R\} & C$\rightarrow$\{A,\:P,\:R\} & P$\rightarrow$\{C,\:A,\:R\} & R$\rightarrow$\{C,\:P,\:A\} & { \# \textbf{Avg.}}\\ \midrule

FedLoA &  77.52 & 83.98 & 74.35 & 74.72 & 77.64\\ 
\cellcolor{cyan!8}{\textbf{FedSCAl}} &  \cellcolor{cyan!8}{\textbf{80.47}} & \cellcolor{cyan!8}{\textbf{86.47}}
& \cellcolor{cyan!8}{\textbf{77.15}} & \cellcolor{cyan!8}{\textbf{76.50}} & \cellcolor{cyan!8}{\textbf{80.15}}\\ \bottomrule

\end{tabular}
\begin{tablenotes}
    \centering
    \item ‘$\rightarrow$’ denotes Source $\rightarrow$ Target; Art (A), ClipArt (C), Product (P), and Real-World (R).
    \end{tablenotes}
\end{threeparttable}}
\vspace{-2pt}
\label{tab:bmd_loa}
\end{table}
\vspace{-14pt}
\begin{table}[htp]
\vspace{-2pt}
\caption{Comparison of FedSCAl and FedLoA using \textbf{UCon} as the underlying LoA method.}
\centering
\vspace{-3pt}
\resizebox{0.47\textwidth}{!}{%
\begin{threeparttable}
\begin{tabular}{l|c|c|c|c|c}
\toprule
Method & A$\rightarrow$\{C,\:P,\:R\} & C$\rightarrow$\{A,\:P,\:R\} & P$\rightarrow$\{C,\:A,\:R\} & R$\rightarrow$\{C,\:P,\:A\} & { \# \textbf{Avg.}}\\ \midrule

FedLoA &  72.00 & 78.55 & 67.25 & 71.85 & 72.41\\ 
\cellcolor{cyan!8}{\textbf{FedSCAl}} &  \cellcolor{cyan!8}{\textbf{73.93}} & \cellcolor{cyan!8}{\textbf{81.77}} & \cellcolor{cyan!8}{\textbf{71.77}} & \cellcolor{cyan!8}{\textbf{74.58}} & \cellcolor{cyan!8}{\textbf{75.51}}\\ \bottomrule

\end{tabular}
\begin{tablenotes}
    \centering
    \item ‘$\rightarrow$’ denotes Source $\rightarrow$ Target; Art (A), ClipArt (C), Product (P), and Real-World (R).
    \end{tablenotes}
\end{threeparttable}}
\vspace{-8pt}
\label{tab:ucon_loa}
\end{table}

Further details on the computation/communication is provided in the Sec. 9 of the supplementary. 

\vspace{-3.0pt}
\section{Limitations and Conclusion}
\vspace{-3.0pt}
Our work introduces FedSCAl, a novel framework for federated learning (FL) designed to address the classification task under the FFreeDA setup. We observe that existing baselines such as FedLoA can suffer from client-drift issues in this setting, primarily due to domain shifts and lack of supervision. To tackle this, FedSCAl incorporates a Server and Client Alignment (SCAl) mechanism, which enables clients to align with the server’s pre-trained model while preserving local adaptation. In doing so, FedSCAl not only mitigates client drift but also presents a clustering-free alternative that addresses the limitations of imperfect clustering in prior methods. The effectiveness of our framework is demonstrated through extensive experiments on the Office-Home, DomainNet, and Office-31 datasets, showing consistent and substantial improvements over baseline and state-of-the-art approaches tailored for this setup. A key limitation of our current approach is that it assumes a predefined number of classes across clients. While this assumption simplifies the setup, it may limit applicability in environments where class distributions evolve over time. Addressing this challenge is an important direction for future work.
\textbf{Acknowledgement:} This work is supported by the
P3DX project, seed-funded by the Ministry of Electronics
and Information Technology (MeitY). The authors would
also like to acknowledge compute and travel supports re-
ceived from Kotak-IISc AI-ML Centre (KIAC), IISc and the
PMRF fellowship.

{
    \small
    \bibliographystyle{ieeenat_fullname}
    \bibliography{main}
}
\clearpage

\appendix

\twocolumn[
\begin{center}
    {\LARGE \bf Supplementary Material}
\end{center}
\vspace{2em}
]

\setcounter{page}{1}

\setcounter{section}{0}
\renewcommand{\cftsecleader}{\cftdotfill{\cftdotsep}}
\renewcommand{\contentsname}{Table of Contents}
\renewcommand{\cftaftertoctitle}{\vspace{-8pt}\par\noindent\rule{\linewidth}{1pt}\vspace{-5pt}\par}

\renewcommand{\cftpnumalign}{l}
\setlength{\cftsecindent}{10pt}
\setlength{\cftsubsecindent}{30pt}
\setlength{\cftsecnumwidth}{17pt}

\setcounter{figure}{0}
\setcounter{table}{0}
\renewcommand{\thesection}{\arabic{section}}


\section{Notations and Terminology}
\begin{itemize}
 \setlength\itemsep{0.2em}
\item $D^s$ denotes the source domain.
\item $g$ denotes the feature extractor.
\item $h$ denotes the classifier.
\item $\mathbb{P}_{s}(x,y)$ represents the source distribution.
\item $\mathbb{P}_{m}(x,y)$ represents the $m^{th}$ domain distribution.
\item $\delta(\cdot)$ is the softmax operator and $\delta_{oh}(\cdot)$ returns the one hot output after the softmax operation. 
\item $\delta_{max}(\cdot)$ denotes the maximum valued element after applying the softmax operator, $\delta(\cdot)$.
\item $\mathbf{w}$ denotes the model parameters.
\item $\mathbf{\tau}$ denotes the adaptive threshold.
\item  $\mathbb{I}_{a > \tau}$ denotes indicator function, which outputs $1$ if $a > \tau$ else $0$.
\item $L_{k}^l$ denotes the client alignment loss for client $k$.
\item $L_{k}^g$ denotes the server alignment loss for client $k$.
\item $L_{k}^{SCAl}$ denotes the server and client alignment loss for client $k$.
\item $L_{k}^{LoA}$ denotes the LoA loss for client $k$.
\item FedLoA is the LoA methods implemented in a federated fashion. 
\item LoA is the pseudo-labelling strategy implemented locally on the clients.
\item FedSCAl is our proposed framework for solving classification tasks in the FFreeDA setup. 
\item $pAcc\:(\text{FedSCAl})$ gives the pseudo-label accuracy when SCAl loss is used along with FedLoA methods. 
\item $pAcc\:(\text{FedLoA})$ is the pseudo-label accuracy of FedLoA methods. 
\item $\Delta_{pAcc}$ is difference in the pseudo-label accuracy of $pAcc\:(\text{FedSCAl})$ and $pAcc\:(\text{FedLoA})$. 

\end{itemize}

\section{Algorithm and Computation Details}

\begin{algorithm}[!h]
 \KwIn{ $\mathbf{w}_s$: pre-trained model on source domain} 
 \KwOut{$\mathbf{w}^{T}$: the final aggregated server model}
 \SetKwInOut{Hyperparameters}{Hyperparameters}
 \Hyperparameters{ $\lambda_{l}$, $\lambda_{g}$, $T$, $E$: local alignment loss weight, global alignment loss weight, communication rounds, local epochs}
 Initialize: server model $\mathbf{w}^0$ $\leftarrow$ $\mathbf{w}_s$\;
 \BlankLine
  \SetKwFunction{cumf}{\textbf{ClientUpdateMeanFeature}}
  \SetKwFunction{CU}{ClientUpdate}
  \SetKwFunction{SA}{ServerAggregate}
   \SetKwFunction{OS}{Optimize the pseudo labeling strategy}
  \SetKwProg{Fn}{Function}{:}{}
  \caption{FedSCAl}
  \label{alg:FedLoA}
  \For{$r \leftarrow 1$ \KwTo $T$}{

    Sever samples $Q$ clients with indices $p_1,\:p_2, \dots ,p_Q$\;
    \For{$m \leftarrow 1$ \KwTo $Q$}{
        server sends model $\mathbf{w}^{r-1}$ to client $p_m$\; 
        $w_{p_m}^r$= \CU{$\mathbf{w}^{r-1}$, $p_m$} \;
       
    }
    $\mathbf{w}^{r}$ = \SA{$\mathbf{w}_{p_1}^{r},...,\mathbf{w}_{p_Q}^{r}$}, 

  }
  \Fn{\CU{$\mathbf{w}^{r-1}$,$p_m$}}{
       set $\mathbf{w}_{k,0} = \mathbf{w}^{r-1}$ \;
       \For{$l \leftarrow 1$ \KwTo $E$}{
       $\mathbf{w}_{k,l} = \mathbf{w}_{k,l-1} - \nabla{l_k}(\mathbf{w}_{k,l-1})$, where
       $l_k(\mathbf{w}_{k,l-1})$ is 
       from
       Eq.~\ref{client_loss} for FedSCAl \; 
    }
    set  $\mathbf{w}_{p_m}^r = \mathbf{w}_{k,E}$\;
  \KwRet $\mathbf{w}_{p_m}^r$\;
  }
  \Fn{\SA{$\mathbf{w}_{p_1}^{r},...,\mathbf{w}_{p_Q}^{r}$}}{
 $\mathbf{w}^{r} = \frac{1} {Q} \sum_{i=1}^{Q} \mathbf{w}_{p_i}^r$
   }
\end{algorithm}

\subsection{Algorithm}


The total client loss $l_k(\mathbf{w})$ for client $k$ is given in Eq.~\ref{client_loss}.

\begin{align}
L_k^{SCAl} &= \lambda_l L_k^l + \lambda_g L_k^g \\
l_k(\mathbf{w}) &= L_k^{LoA} + L_k^{SCAl}  
\label{client_loss}
\end{align}

Here, $L_k^{LoA}$ denotes the LoA loss for client $k$, and the alignment loss is given by $L_k^{SCAl}$. The parameters $\lambda_l$ and $\lambda_g$ represent the client and server alignment loss weights, respectively. A detailed discussion of the algorithm used to implement our SCAl loss with the pseudo-labeling strategy is provided in Algorithm~\ref{alg:FedLoA}.

\subsection{Setup Use Cases} 
When federated learning (FL) is implemented across multiple surveillance camera sites, the requirement to manually tag each image becomes a significant administrative burden. Although servers managed by organizations often have access to labelled datasets, the sensitive biometric details—such as facial features—embedded in those images lead to privacy complications when using them directly for FL, thus necessitating the use of a pre-trained server model to initialize the federated training. Furthermore, disparities in conditions like lighting and background across various camera locations exemplify the challenges posed by differing domain characteristics.

\begin{table*}[t]
\renewcommand{\arraystretch}{1.2}
\caption{Comparison of {FedLoA} and {FedSCAl} on \textbf{Office-Home} dataset, when BMD is used as underlying LoA method.}
\centering
\resizebox{\textwidth}{!}{
\begin{tabular}{l|ccc|c|ccc|c|ccc|c|ccc|c}
\toprule
\multirow{2}{*}{Method} & \multicolumn{4}{c|}{ \makecell{Initial Sever Model is pre-trained \\on \textbf{\texttt{Art}}}} & \multicolumn{4}{c|}{\makecell{ Initial Sever Model is pre-trained \\on \textbf{\texttt{Clipart}}}} & \multicolumn{4}{c|}{ \makecell{Initial Sever Model is pre-trained \\on \textbf{\texttt{Product}}}} & \multicolumn{4}{c}{ \makecell{Initial Sever Model is pre-trained \\on \textbf{\texttt{Real}}}} \\

& \textbf{Clipart} & \textbf{Product} & \textbf{Real} & \textbf{Avg.} & \textbf{Art} & \textbf{Product} & \textbf{Real} & \textbf{Avg.} & \textbf{Art} & \textbf{Clipart} & \textbf{Real} & \textbf{Avg.} & \textbf{Art} & \textbf{Clipart} & \textbf{Product} & \textbf{Avg.} \\ \midrule
FedLoA & 58.74 & 86.06 & 87.77 & 77.52 & 80.2 & 85.05 & 86.7 & 83.98 & 78 & 57.59 & 87.45 & 74.35 & 78.42 & 57.93 & 87.8 & 74.72 \\ 
 \cellcolor{cyan!8}{\textbf{FedSCAl} (Ours)} & \cellcolor{cyan!8}{\textbf{63.84}} & 
\cellcolor{cyan!8}{\textbf{88.39}} & \cellcolor{cyan!8}{\textbf{89.19}} & \cellcolor{cyan!8}{\textbf{80.47}} & \cellcolor{cyan!8}{\textbf{81.62}} & \cellcolor{cyan!8}{\textbf{88.88}} & \cellcolor{cyan!8}{\textbf{88.91}} & \cellcolor{cyan!8}{\textbf{86.47}} & \cellcolor{cyan!8}{\textbf{80.1}} & \cellcolor{cyan!8}{\textbf{62.98}} & \cellcolor{cyan!8}{\textbf{88.37}} & \cellcolor{cyan!8}{\textbf{77.15}} & \cellcolor{cyan!8}{\textbf{79.51}} & \cellcolor{cyan!8}{\textbf{60.81}} & \cellcolor{cyan!8}{\textbf{89.19}} & \cellcolor{cyan!8}{\textbf{76.5}} \\

\bottomrule
\end{tabular}
}
\label{tab:office_home_bmd}
\end{table*}

\begin{table*}[t]
\renewcommand{\arraystretch}{1.45}
\caption{Results on \textbf{Domain-Net} Dataset, when the underlying LoA method used is \textbf{BMD}. We report the accuracy (\%) for each target domain and the average of all the domains for each pre-trained model. It is evident that adding our SCAl loss boosts the performance across all the domains.}

\centering
\scalebox{0.7}
{
\begin{tabular}{l|ccccc|c|ccccc|c}
\toprule
\multirow{2}{*}{Method} & \multicolumn{6}{c|}{(a) Initial Sever Model is pre-trained on \textbf{\texttt{Clipart}}} & \multicolumn{6}{c}{(b) Initial Sever Model is pre-trained on \textbf{\texttt{Infograph}}} \\

& \textbf{Infograph} & \textbf{Painting} & \textbf{Quickdraw} & \textbf{Real} & \textbf{Sketch} & \textbf{Avg.} & \textbf{Clipart} & \textbf{Painting} & \textbf{Quickdraw} & \textbf{Real} & \textbf{Sketch} & \textbf{Avg.} \\ \midrule

FedLoA & 56.96 & 92.48 & 73.24 & 96.28 & 87.28 & 81.24& 81.63 & 91.10 & 66.93 & 95.82 & 82.83 & 83.66 \\ 

\cellcolor{cyan!8}{\textbf{FedSCAl} (Ours)} & \cellcolor{cyan!8}{\textbf{58.19}} & \cellcolor{cyan!8}{\textbf{94.89}} & \cellcolor{cyan!8}{\textbf{76.68}} & \cellcolor{cyan!8}{\textbf{96.60}} & \cellcolor{cyan!8}{\textbf{91.94}} & \cellcolor{cyan!8}{\textbf{83.66}} & 
\cellcolor{cyan!8}{\textbf{86.38	}} & \cellcolor{cyan!8}{\textbf{94.62	}} & \cellcolor{cyan!8}{\textbf{76.44}} & \cellcolor{cyan!8}{\textbf{96.78}} & \cellcolor{cyan!8}{\textbf{89.37}} & \cellcolor{cyan!8}{\textbf{88.72}}\\ \bottomrule

\multirow{2}{*}{Method} & \multicolumn{6}{c|}{(c) Initial Sever Model is pre-trained on \textbf{\texttt{Painting}}} & \multicolumn{6}{c}{(d) Initial Sever Model is pre-trained on \textbf{\texttt{Quickdraw}}} \\

& \textbf{Clipart} & \textbf{Infograph} & \textbf{Quickdraw} & \textbf{Real} & \textbf{Sketch} & \textbf{Avg.} & \textbf{Clipart} & \textbf{Infograph} & \textbf{Painting} & \textbf{Real} & \textbf{Sketch}  & \textbf{Avg.} \\ \midrule
FedLoA & 80.24 & 52.09 & 66.97 & 95.68 & 80.09 & 75.01 & 61.51 & 38.00 & 58.35 & 62.10 & 60.72 & 56.13\\ 

\cellcolor{cyan!8}{\textbf{FedSCAl} (Ours)} & \cellcolor{cyan!8}{\textbf{87.03}} & \cellcolor{cyan!8}{\textbf{56.59}} & \cellcolor{cyan!8}{\textbf{77.82}} & \cellcolor{cyan!8}{\textbf{96.60	}} & \cellcolor{cyan!8}{\textbf{88.83	}} & \cellcolor{cyan!8}{\textbf{81.38}} & \cellcolor{cyan!8}{\textbf{71.57	}} & \cellcolor{cyan!8}{\textbf{44.40}} & \cellcolor{cyan!8}{\textbf{74.65}} & \cellcolor{cyan!8}{\textbf{75.04}}  & \cellcolor{cyan!8}{\textbf{72.21}} & \cellcolor{cyan!8}{\textbf{67.57}}\\ \bottomrule

\multirow{2}{*}{Method} & \multicolumn{6}{c|}{(e) Initial Sever Model is pre-trained on \textbf{\texttt{Real}}} & \multicolumn{6}{c}{(f) Initial Sever Model is pre-trained on \textbf{\texttt{Sketch}}} \\ 

& \textbf{Clipart} & \textbf{Painting} & \textbf{Quickdraw} & \textbf{Infograph} & \textbf{Sketch} & \textbf{Avg.} & \textbf{Clipart} & \textbf{Painting} & \textbf{Quickdraw} & \textbf{Real} & \textbf{Infograph} & \textbf{Avg.}\\ \midrule
FedLoA & 71.47 & 87.12 & 52.30 & 47.79 & 76.82 & 67.10 & 82.67 & 90.44 & 75.48 & 96.39 & 53.61 & 79.72\\ 

\cellcolor{cyan!8}{\textbf{FedSCAl} (Ours)} & \cellcolor{cyan!8}{\textbf{80.14}} & \cellcolor{cyan!8}{\textbf{	91.51	}} & \cellcolor{cyan!8}{\textbf{68.82}} & \cellcolor{cyan!8}{\textbf{	52.97	}}  & \cellcolor{cyan!8}{\textbf{84.33}} & \cellcolor{cyan!8}{\textbf{	75.55}} &
\cellcolor{cyan!8}{\textbf{88.25	}} & \cellcolor{cyan!8}{\textbf{94.53}} & \cellcolor{cyan!8}{\textbf{	79.05}} & \cellcolor{cyan!8}{\textbf{	96.65	}} & \cellcolor{cyan!8}{\textbf{57.20}} & \cellcolor{cyan!8}{\textbf{	83.14}} \\ \bottomrule

\end{tabular}
}
\label{tab:domain_net_bmd}
\end{table*}

\section{Alternative peseudo-labeling methods}

\paragraph{BMD~\cite{qu2022bmd}}
We leverage the Inter-class Balanced Prototype and Dynamic Pseudo Label findings from BMD.
Class-biased strategies tend to aggregate biased data instances from easy-transfer classes, potentially generating noisy labels for hard data. To address this, the authors propose a global inter-class balanced sampling strategy, formulated as a multiple-instance learning problem. In this approach, each data instance in the target domain is represented by a feature vector and a classification result. The top-M most likely instances are aggregated to construct class-balanced feature prototypes, facilitating pseudo-label assignment as shown below
\begin{equation}
M_j = \underset{\substack{x_t \in X_t \\ |M_j|=M}}{\arg\max} \; \delta_k\left(f^t(x_t)\right)
\end{equation}

\begin{equation}
    c_j = \frac{1}{M}\sum_{i \in M_j} \hat g^t(x_t^{i})
\end{equation}
\begin{equation}    
    \hat {y}_t = \arg \min _j D_f(\hat {g}^t(x_t), c_j)
\end{equation}
where $M = max\{1, \lfloor \frac{n_t}{r*K}\rfloor \}$, r is a hyperparameter denoting the top-M selection ratio, and J is the number of object classes in the target domain.
To improve performance, a dynamic pseudo-labeling approach is deployed. At each epoch's start, feature prototypes and corresponding pseudo labels are globally updated. During iteration steps, feature prototypes are adjusted using an exponential moving average of cluster centroids. Dynamic pseudo labels are computed based on instance-feature prototype similarities, normalized across classes. A robust symmetric cross-entropy loss is adopted over standard cross-entropy. However, dynamic pseudo labels may overlook domain shifts, leading to less informative prototypes. To mitigate this, the static pseudo label-based self-training loss is combined with dynamic loss, balanced by hyper-parameters $\alpha$ and $\beta$. The dynamic pseudo labels $\hat{y_t}^{d}$ and the feature prototypes are updated as follows
\begin{equation}
    \hat{y_t}^{d} = \frac{exp(\hat {g}^t(x_t), c_j)}{\sum_{k=1}^K exp(\hat {g}^t(x_t), c_j)}
\end{equation}
\begin{equation}
    p_j({x_t}^n)  = \frac{exp(\hat {g}^t({x_t}^n), c_j)}{\sum_{j=1}^J exp(\hat {g}^t({x_t}^n), c_j)}
\end{equation}
\begin{equation}
    \hat{c_j} = \frac{\sum_{n=1}^N \hat {g}^t({x_t}^n)*p_j({x_t}^n)}{\sum_{n=1}^N p_j({x_t}^n)}
\end{equation}
\begin{equation}
    c_j = \lambda{c_j}+(1-\lambda)\hat{c_j}
\end{equation}
\begin{equation}
L_{st} = -\frac{1}{N} \sum_{i=1}^{N} \sum_{k=1}^{K} \mathbb{I}[k = \hat{y}_t] \log \delta_k(f^t(x_t^i))
\end{equation}

where $p_j^i(x_t)$ denotes the similarity of instance $x_t$ with existing feature prototypes, $\hat{c_j}$
represents the feature prototype of class k calculated with the current training mini-batch, and $\lambda$ is the momentum coefficient of EMA. Thus $L
^{bmd}$ becomes our $L_k^{LoA}$\:(Eq.~\ref{client_loss})
\begin{equation}
\begin{split}
  L^{dyn} &= -\frac{1}{N}\sum_{i=1}^{N}\sum_{j=1}^{J}\hat{y}_{t,j}^d \log( \delta _j({f^t}({x_t}^{i}))) - \\
        & \ \frac{1}{N}\sum_{i=1}^{N}\sum_{j=1}^{J}\delta _j({f^t}({x_t}^{i})) \log(\hat{y}_{t,j}^d)
\end{split}
\end{equation}
\begin{equation}
L^{LoA} = \alpha*L^{st} + \beta*L^{dyn}
\label{bmd_eq}
\end{equation}

\paragraph{UCon-SFDA~\cite{xu2025revisiting}}
The UCon‑SFDA method extends the conventional contrastive loss for source‑free domain adaptation by introducing uncertainty control mechanisms that target both negative and positive sample selection.  
\begin{equation}
L_{CL} = L_{CL}^{+} \;+\;\lambda_{CL}^{-}\,L_{CL}^{-}
\end{equation}
Here, 
\begin{equation}
L_{CL}^{+} = -\frac{1}{N_T}\sum_{i=1}^{N}\sum_{x_i^+\in C_i} S_\theta(x_i^+;x_i),
\end{equation}
\begin{equation}
L_{CL}^{-} = \frac{1}{N}\sum_{i=1}^{N}\sum_{x_i^-\in B\setminus\{x_i\}} S_\theta(x_i^-;x_i),
\end{equation}
and \(S_\theta(u;v)=\langle f(u;\theta),f(v;\theta)\rangle\) denotes the dot‑product similarity between feature embeddings.  
To mitigate false negatives, a dispersion control term is added:  
\begin{equation}
L_{DC} = -\frac{1}{N}\sum_{i=1}^{N} d_\theta\bigl(\mathrm{AUG}(x_i),x_i\bigr),
\end{equation}
where $\mathrm{AUG}(\cdot)$  denotes a stochastic data augmentation and 
$d_\theta(u,v)=\langle f(u;\theta),\log f(v;\theta)\rangle$
The negative component of the uncertainty‑controlled loss becomes  

\begin{equation}
L_{UCon}^{-} = \lambda_{CL}^{-} L_{CL}^{-}  + \lambda_{DC} L_{DC}
\end{equation}

with $\lambda_{DC}$ as the dispersion weight.  
On the positive side, an uncertain set \(\mathcal U\) is defined by the confidence ratio criterion \(p_{(1)}/p_{(2)} \le \tau\), where \(p = f(x;\theta)\) and \(p_{(k)}\) is its \(k\)‑th largest entry.  For each \(x_i \in \mathcal U\), a partial‑label set \(\mathcal Y^{PL}_{x_i}\) of size \(K_{PL}\) is formed from its top‑\(K_{PL}\) historical predictions.  The relaxed positive loss then reads  
\begin{dmath}
L_{UCon}^{+} = L_{CL}^{+} + \lambda_{PL} \frac{1}{N} \sum_{i=1}^{N} \sum_{y \in \mathcal{Y}^{PL}_{x_i}} \mathbf{1}[x_i \in \mathcal{U}] \, \ell_{\mathrm{CE}}\bigl(y,\, f(x_i; \theta)\bigr)
\end{dmath}
where \(\ell_{\mathrm{CE}}(\cdot,\cdot)\) is the cross‑entropy loss and \(\lambda_{PL}\) the partial‑label weight.  Finally, the overall adaptation objective is  
\begin{equation}
L^{LoA} = L_{UCon}^{+} \;+\; L_{UCon}^{-},
\end{equation}
balancing dispersion and partial‑label contributions.

\begin{table*}[t]
\renewcommand{\arraystretch}{1.2}
\caption{Comparison of {FedLoA} and {FedSCAl} on \textbf{Office-Home} dataset, when UCon is used as underlying LoA method.}
\centering
\resizebox{\textwidth}{!}{
\begin{tabular}{l|ccc|c|ccc|c|ccc|c|ccc|c}
\toprule
\multirow{2}{*}{Method} & \multicolumn{4}{c|}{ \makecell{Initial Sever Model is pre-trained \\on \textbf{\texttt{Art}}}} & \multicolumn{4}{c|}{\makecell{ Initial Sever Model is pre-trained \\on \textbf{\texttt{Clipart}}}} & \multicolumn{4}{c|}{ \makecell{Initial Sever Model is pre-trained \\on \textbf{\texttt{Product}}}} & \multicolumn{4}{c}{ \makecell{Initial Sever Model is pre-trained \\on \textbf{\texttt{Real}}}} \\

& \textbf{Clipart} & \textbf{Product} & \textbf{Real} & \textbf{Avg.} & \textbf{Art} & \textbf{Product} & \textbf{Real} & \textbf{Avg.} & \textbf{Art} & \textbf{Clipart} & \textbf{Real} & \textbf{Avg.} & \textbf{Art} & \textbf{Clipart} & \textbf{Product} & \textbf{Avg.} \\ \midrule

FedLoA & 53.44	& 80.23 &	82.35 &	72.01 &	74.4 &	80.51	&80.76&	78.56	&70.09	&50.16	&81.52&	67.26&	75.92&	55&	84.65&	71.86 \\ 

\cellcolor{cyan!8}{\textbf{FedSCAl} (Ours)} & \cellcolor{cyan!8}{\textbf{57.76}} &  
\cellcolor{cyan!8}{\textbf{	81.34}} & \cellcolor{cyan!8}{\textbf{	82.71}} & \cellcolor{cyan!8}{\textbf{	73.94}} & \cellcolor{cyan!8}{\textbf{77.17	}} & \cellcolor{cyan!8}{\textbf{82.99	}} & \cellcolor{cyan!8}{\textbf{85.16}} & \cellcolor{cyan!8}{\textbf{	81.77}} & \cellcolor{cyan!8}{\textbf{75.9}} & \cellcolor{cyan!8}{\textbf{	55.46}} & \cellcolor{cyan!8}{\textbf{	83.96}} & \cellcolor{cyan!8}{\textbf{	71.77}} & \cellcolor{cyan!8}{\textbf{77.6}} & \cellcolor{cyan!8}{\textbf{	59.34}} & \cellcolor{cyan!8}{\textbf{	86.82}} & \cellcolor{cyan!8}{\textbf{	74.59}} \\

\bottomrule
\end{tabular}
}
\label{tab:office_home_ucon}
\end{table*}

\begin{table*}[htp]
\renewcommand{\arraystretch}{1.45}
\caption{Results on \textbf{Domain-Net} Dataset, when the underlying LoA method used is \textbf{UCon}. We report the accuracy (\%) for each target domain and the average of all the domains for each pre-trained model. Clearly, adding our SCAL loss boosts the performance.}

\centering
\scalebox{0.7}
{
\begin{tabular}{l|ccccc|c|ccccc|c}
\toprule
\multirow{2}{*}{Method} & \multicolumn{6}{c|}{(a) Initial Sever Model is pre-trained on \textbf{\texttt{Clipart}}} & \multicolumn{6}{c}{(b) Initial Sever Model is pre-trained on \textbf{\texttt{Infograph}}} \\

& \textbf{Infograph} & \textbf{Painting} & \textbf{Quickdraw} & \textbf{Real} & \textbf{Sketch} & \textbf{Avg.} & \textbf{Clipart} & \textbf{Painting} & \textbf{Quickdraw} & \textbf{Real} & \textbf{Sketch} & \textbf{Avg.} \\ \midrule
FedLoA & 56.94&	93.31	&67.76&	96.33	&87.06&	80.28&	80.63&	91.86&	66.03&	95.86&	81.51	&83.18 \\ 

\cellcolor{cyan!8}{\textbf{FedSCAl} (Ours)} & \cellcolor{cyan!8}{\textbf{60.07	}} & \cellcolor{cyan!8}{\textbf{94.42}} & \cellcolor{cyan!8}{\textbf{	74.14	}} & \cellcolor{cyan!8}{\textbf{96.51	}} & \cellcolor{cyan!8}{\textbf{90.22	}} & \cellcolor{cyan!8}{\textbf{83.08}} & 
\cellcolor{cyan!8}{\textbf{85.63	}} & \cellcolor{cyan!8}{\textbf{93.56}} & \cellcolor{cyan!8}{\textbf{	76	}} & \cellcolor{cyan!8}{\textbf{96.58}} & \cellcolor{cyan!8}{\textbf{	86.24}} & \cellcolor{cyan!8}{\textbf{	87.6	}}\\ \bottomrule

\multirow{2}{*}{Method} & \multicolumn{6}{c|}{(c) Initial Sever Model is pre-trained on \textbf{\texttt{Painting}}} & \multicolumn{6}{c}{(d) Initial Sever Model is pre-trained on \textbf{\texttt{Quickdraw}}} \\

& \textbf{Clipart} & \textbf{Infograph} & \textbf{Quickdraw} & \textbf{Real} & \textbf{Sketch} & \textbf{Avg.} & \textbf{Clipart} & \textbf{Infograph} & \textbf{Painting} & \textbf{Real} & \textbf{Sketch}  & \textbf{Avg.} \\ \midrule

FedLoA &81.02&	53.85&	61.36	&96.12&	83.27&	75.12&	47.61&	37.89&	53.75	&57.07&	48.14&	48.89\\ 

\cellcolor{cyan!8}{\textbf{FedSCAl} (Ours)} & \cellcolor{cyan!8}{\textbf{85.23	}} & \cellcolor{cyan!8}{\textbf{58.58}} & \cellcolor{cyan!8}{\textbf{	68.63	}} & \cellcolor{cyan!8}{\textbf{96.32}} & \cellcolor{cyan!8}{\textbf{	86.24}} & \cellcolor{cyan!8}{\textbf{	79	}} & \cellcolor{cyan!8}{\textbf{68.4	}} & \cellcolor{cyan!8}{\textbf{50.86}} & \cellcolor{cyan!8}{\textbf{	70.68}} & \cellcolor{cyan!8}{\textbf{	71.45}}  & \cellcolor{cyan!8}{\textbf{	65.46	}} & \cellcolor{cyan!8}{\textbf{68.37}}\\ \bottomrule

\multirow{2}{*}{Method} & \multicolumn{6}{c|}{(e) Initial Sever Model is pre-trained on \textbf{\texttt{Real}}} & \multicolumn{6}{c}{(f) Initial Sever Model is pre-trained on \textbf{\texttt{Sketch}}} \\ 

& \textbf{Clipart} & \textbf{Painting} & \textbf{Quickdraw} & \textbf{Infograph} & \textbf{Sketch} & \textbf{Avg.} & \textbf{Clipart} & \textbf{Painting} & \textbf{Quickdraw} & \textbf{Real} & \textbf{Infograph} & \textbf{Avg.}\\ \midrule

FedLoA & 81.53&	93.08&	64.63&	56.5&	85.49	&76.25&	84.25&	93.92&	70.98&	96.33&	57.98&	80.69\\ 

\cellcolor{cyan!8}{\textbf{FedSCAl} (Ours)} & \cellcolor{cyan!8}{\textbf{85.48	}} & \cellcolor{cyan!8}{\textbf{94.21}} & \cellcolor{cyan!8}{\textbf{	73.85	}} & \cellcolor{cyan!8}{\textbf{59.86}}  & \cellcolor{cyan!8}{\textbf{	88.86	}} & \cellcolor{cyan!8}{\textbf{80.45}} &
\cellcolor{cyan!8}{\textbf{	88.77	}} & \cellcolor{cyan!8}{\textbf{	94.78	}} & \cellcolor{cyan!8}{\textbf{	76.04	}} & \cellcolor{cyan!8}{\textbf{	96.5	}} & \cellcolor{cyan!8}{\textbf{60.21}} & \cellcolor{cyan!8}{\textbf{		83.26	}} \\ \bottomrule

\end{tabular}
}
\label{tab:domain_net_bmd}
\end{table*}

\section{Results with LoA as BMD and UCon}

The two tables~\ref{tab:office_home_bmd} and~\ref{tab:domain_net_bmd} present a comparison between {FedLoA} and the proposed {FedSCAl} method on the Office-Home and {DomainNet} datasets, using {BMD} as the underlying Local Adaptation (LoA) method. In both tables, different initial server model pre-training domains are considered, and performance is evaluated on the remaining target domains. Across all settings, {FedSCAl} consistently outperforms {FedLoA}, demonstrating better generalization to unseen domains. For instance, on Office-Home, when pre-trained on \texttt{Art}, the average accuracy improves from {77.52\%} (FedLoA) to {80.47\%} (FedSCAl), and similar gains are observed across other source domains. Likewise, on the more challenging DomainNet dataset, FedSCAl shows strong improvements in every pre-training scenario for example, when initialized on \texttt{Painting}, the average accuracy increases from {75.01\%} to {81.38\%}, and from {79.72\%} to {83.14\%} when initialized on {Sketch}. These results confirm that incorporating the SCAl loss enhances adaptation and improves performance across diverse and unseen target domains. Similarly the table~\ref{tab:office_home_ucon} compares the performance of {FedLoA} and the proposed {FedSCAl} method on the {Office-Home} dataset, using {UCon} as the underlying Local Adaptation (LoA) method. Results are reported across four different initial server pre-training domains: \texttt{Art}, \texttt{Clipart}, \texttt{Product}, and \texttt{Real}, with accuracy evaluated on the remaining target domains. Across all configurations, {FedSCAl} consistently outperforms {FedLoA}. For example, when the initial model is pre-trained on \texttt{Clipart}, the average accuracy improves from {78.56\%} (FedLoA) to {81.77\%} (FedSCAl), and similarly, for the \texttt{Real} pre-training setting, the average increases from {71.86\%} to {74.59\%}. These results show that integrating the SCAl loss enhances the adaptation quality and leads to improved generalization under UCon-based local training.

\begin{figure}[htp]
         \centering
         \begin{subfigure}[b]{0.3\textwidth}
         \centering
         \includegraphics[width=\textwidth]
         {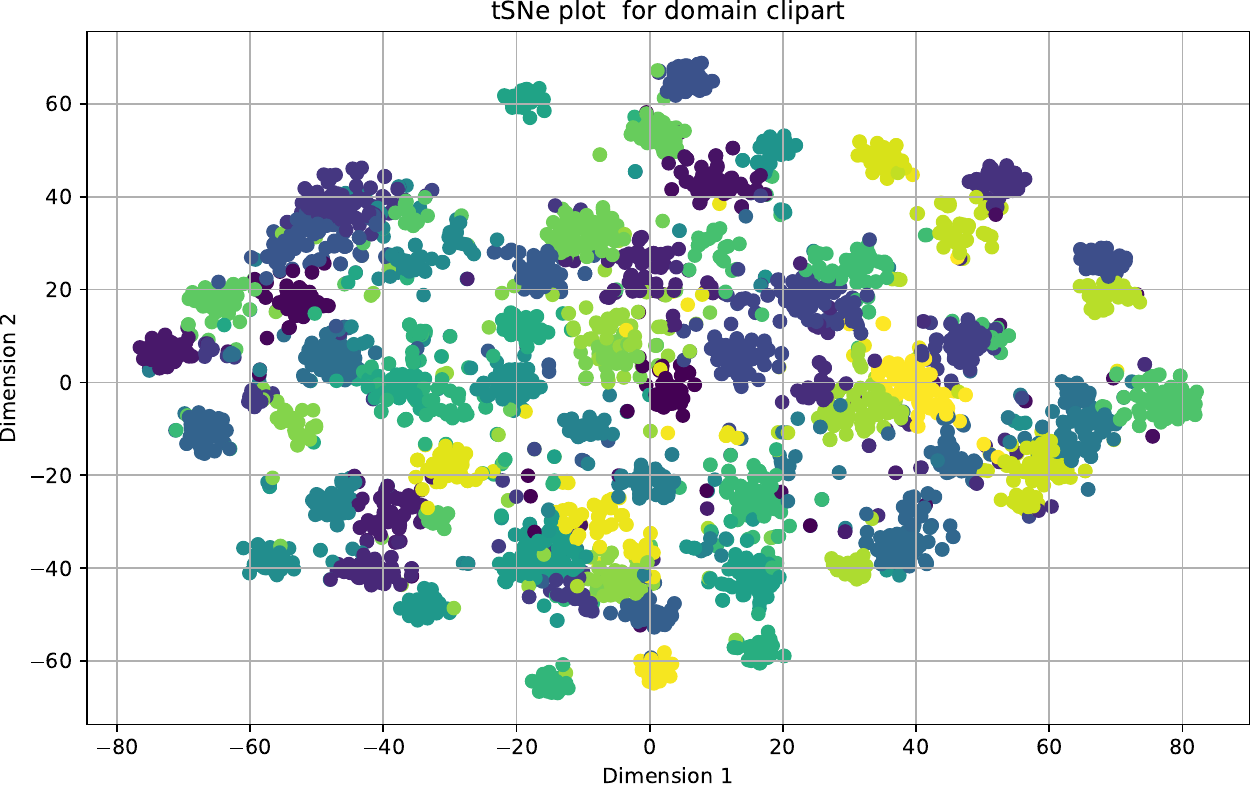}
         \caption{t-SNE plot for FedLoA when the server model is pre-trained 
         on \textbf{\texttt{Art}} and the client has \textbf{\texttt{Clipart}} data.}
         \label{fig:art source}
         \end{subfigure}
         \hfill
         \begin{subfigure}[b]{0.3\textwidth}
         \centering
         \includegraphics[width=\textwidth]
         {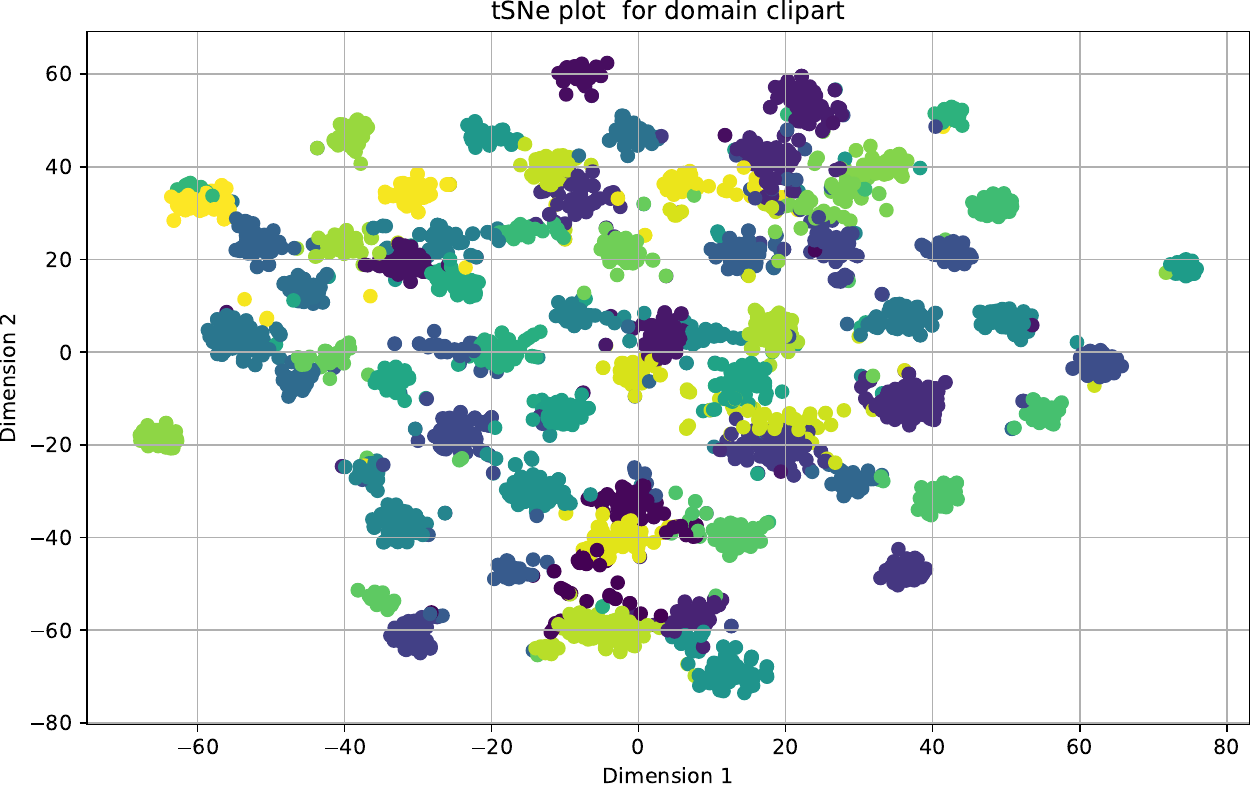}
         \caption{t-SNE plot for FedSCAl  when the server model is pre-trained on \textbf{\texttt{Art}} and the client has \textbf{\texttt{Clipart}} data.}
         \label{fig:art source}
         \end{subfigure}
         \hfill
         \caption{\textbf{(a)} t-SNE plot of representations learned by FedLoA, \textbf{(b)} t-SNE plot for representations learned after adding our proposed loss (FedSCAl). It can be seen that adding our SCAl loss leads to better clustering of the learned representations.}
        
     \label{tsne}
     \vspace{-0.1in}
\end{figure}
\section{T-SNE Visualization of representations}
In Fig.~\ref{tsne}, we visualize the t-SNE embeddings of the learned representations on a client from the \texttt{Clipart} domain, where the initial server model is pre-trained on the \texttt{Art} domain. Under FedLoA, the representations exhibit overlapping and poorly formed class clusters, indicating suboptimal adaptation to the target domain. In contrast, FedSCAl leads to relatively well-separated and compact clusters, demonstrating improved inter-class separability. This highlights that FedSCAl is more effective in aligning the source-pretrained server model with the client's target distribution, particularly in challenging cross-domain scenarios. The improved clustering structure validates the effectiveness of SCAl loss in enhancing federated adaptation and feature discriminability.


\begin{table*}[t]
\caption{Results on \textbf{Office-31} Dataset, while the initial server pre-trained model is trained on one of the three domains (Amazon, Webcam, and DSLR), and the clients are distributed with the remaining two domains such that each client contains a subset of exactly one Domain.}
\centering
\scalebox{0.85}{
\begin{tabular}{l|cc|c|cc|c|cc|c}
\toprule
\multirow{2}{*}{Method} & \multicolumn{3}{c|}{{(a) Initial Server Model: \textbf{Amazon}}} & \multicolumn{3}{c|}{{(b) Initial Server Model: \textbf{DSLR}}} & \multicolumn{3}{c}{{(c) Initial Server Model: \textbf{Webcam}}} \\
& \textbf{DSLR} & \textbf{Webcam} & \textbf{Avg.} & \textbf{Amazon} & \textbf{Webcam} & \textbf{Avg.} & \textbf{Amazon} & \textbf{DSLR} & \textbf{Avg.} \\
\midrule
LoA & 92.97 & 94.21 & 93.59 & 75.61 & 93.96 & 84.79 & 78.83 & 98.93 & 88.61 \\ \midrule
FedLoA & 96.65 & 95.64 & 96.15 & 81.11 & 95.68 & 88.39 & 79.19 & 97.79 & 88.56 \\ \midrule

FedProx & 95.14	& 94.58	& 94.86 & 79.03	& 94.58 & 86.81 & 77.35 & 97.56 & 87.46\\ \midrule
FedMOON & 95.72	& 95.05	& 95.39 & 80.15	& 95.21 & 
 87.68 & 78.79 & 97.61 & 88.20\\ \midrule
FedWCA & 95.65 & 90.9 & 93.28 & 75.66 & 97.78 & 86.72 & 75.88 & 99.73 & 87.81 \\ \midrule
LADD & 84.54 & 85.03 & 84.78 & 63.9 & 94.72 & 79.31 & 70.6 & 99.4 & 84.99 \\ \midrule
Dual Adapt & 88.86 & 87.42 & 88.14 & 69.08 & 97.02 & 83.05 & 73.81 & 99.35 & 86.58 \\ \midrule
\cellcolor{cyan!8} \textbf{FedSCAl} (Ours) & \cellcolor{cyan!8}{96.65} & \cellcolor{cyan!8}{96.27} & \cellcolor{cyan!8}\textbf{96.46} & \cellcolor{cyan!8}{81.69} & \cellcolor{cyan!8}96.81 & \cellcolor{cyan!8}\textbf{89.25} & \cellcolor{cyan!8}{82.87} & \cellcolor{cyan!8}98.66 & \cellcolor{cyan!8}\textbf{90.76} \\
\bottomrule
\end{tabular}
}
\label{tab1:office31}
\end{table*}
\section{Additional Experiments }
As mentioned in the main paper, due to space constraints we provide the Office-31 results in Table~\ref{tab1:office31}. It reports the results on the Office-31 dataset, where the initial server model is pre-trained on one of the three domains (Amazon, DSLR, or Webcam), and clients are drawn from the remaining two domains, with each client containing data from only a single domain. The performance is measured on the held-out client domains and reported individually as well as averaged. The proposed FedSCAl method consistently outperforms all baselines across all settings. For example, when the server is initialized on Amazon, FedSCAl achieves the highest average accuracy of 96.46\%, surpassing methods like FedLoA (96.15\%), FedProx (94.86\%), and FedMOON (95.39\%). Similarly, in the DSLR-pretrained setting, FedSCAl reaches an average of 89.25\%, showing gains over FedLoA (88.39\%) and other competing methods. In the Webcam-pretrained case, FedSCAl achieves the highest average accuracy of 90.76\%, demonstrating its robust cross-domain adaptation ability under varied initialization conditions.

 
It can be seen that the LoA method performs inferior to FedLoA and FedSCAl even though there is no client-drift in this case, the key issue being the training on limited single client data. In the case of FedLoA, the performance is not the best as client-drift limits its performance. FedSCAl performs the best as it minimizes the client-drift across the domains.
It can be seen that FedSCAl outperforms the FedLoA baseline by $2.2\%$ when the server model is pre-trained on Webcam data, and it beats FedLoA by $0.86\%$ when the server model is pre-trained on DSLR data.

\section{Datasets and Hyper-parameter Settings }
Office-31~\cite{10.1007/978-3-642-15561-1_16} dataset contains three domains: Amazon, DSLR, and Webcam. While we consider one domain as a source, the other $2$ domains are distributed among $6$ clients ($3$ each for a particular domain). The participation rate is set to $0.5$.
For Office-Home~\cite{venkateswara2017deep} and Office-31 we set $\lambda_{l}$ and $\lambda_{g}$ to $1$.
For Domain-Net~\cite{Gong2012GeodesicFK} we set $\lambda_{l}$ to $3$ and $\lambda_{g}$ to $0.3$. We follow the protocol of ~\cite{li2021fedbn} for adapting the Domain-Net dataset to the federated setup. 
We use an SGD optimizer for adaptation and a learning rate of $0.03$ for the Office-Home, Office-31, and Domain-Net datasets. A batch size of 64 is used for training at each client for all datasets. Local training epochs for each client are kept at 5 for all datasets. As FedProx and FedMOON are designed for supervised settings, we implement them in the FFreeDA setting with the LoA method as described in Section 3.1 of the main paper.

\section{Analysis of SCAl by varying the $\lambda_l$ and $\lambda_g$}

\begin{figure}[htp]
    \centering     
    \includegraphics[width=0.5\textwidth]{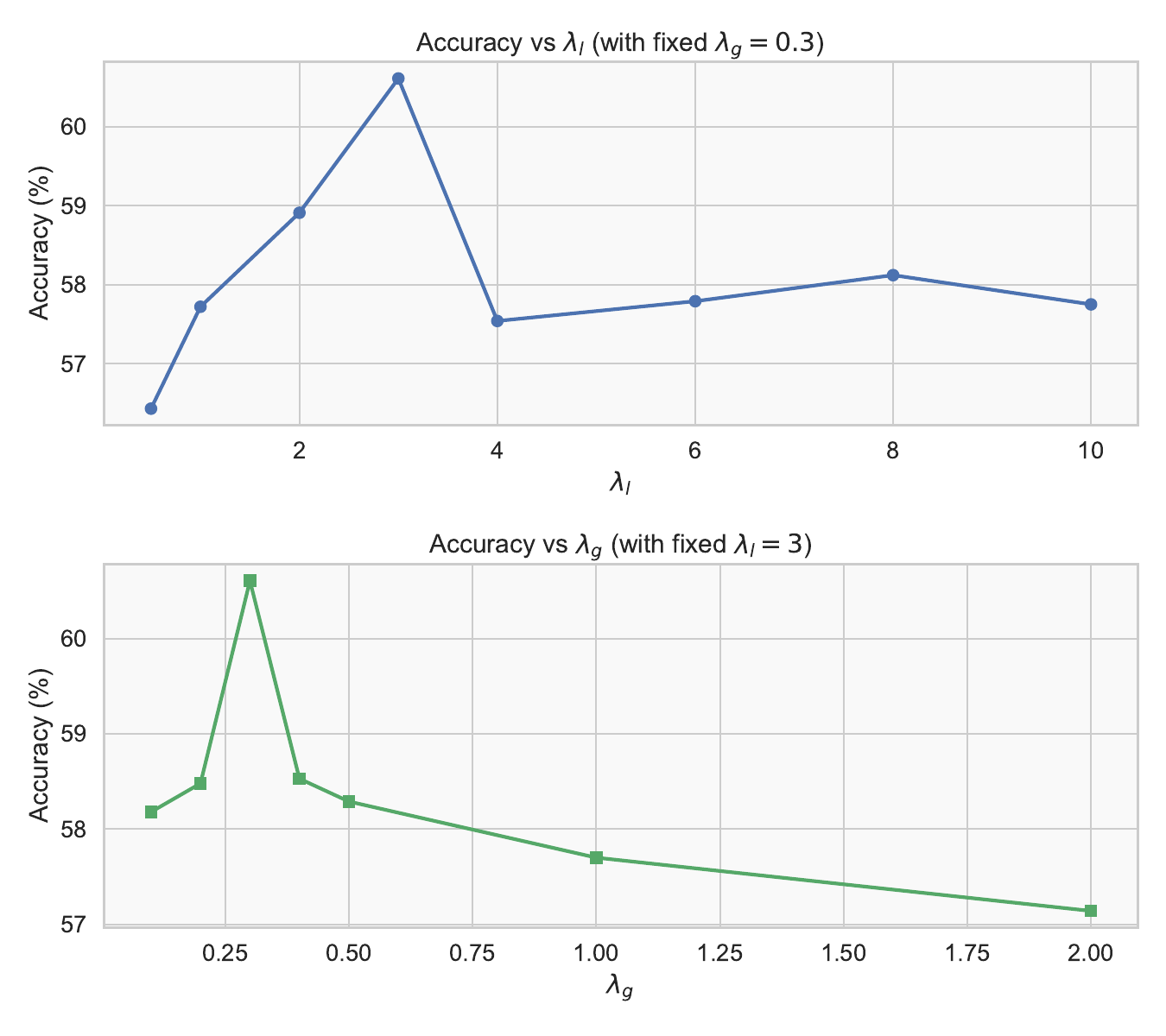}
    \caption{Ablation of SCAl Regularization Weights: Average accuracy across clients while varying $\lambda_l$ and $\lambda_g$.}
    \label{fig:reg_abl}        
\end{figure}

We perform an ablation study on the regularization weights $\lambda_l$ and $\lambda_g$, where the initial server model is pre-trained on the \texttt{Quickdraw} domain, and the clients are distributed with data from the remaining domains. The plots in the Figure~\ref{fig:reg_abl} show the average accuracy across all clients as we vary one regularizer while keeping the other fixed. We observe that while there is some fluctuation in performance, the overall accuracy remains relatively stable across a range of $\lambda_l$ and $\lambda_g$.

\section{Computation and Communication Cost}
If $c$ denotes the Forward Pass Computation Cost (FPCC), and $e$ is the number of local epochs, then the proposed regularizer will have an effective FPCC of $ce$+$ce$+$c$. This effectively incurs only twice the FPCC for client training in each communication round when the local epochs $e$ is larger than $1$. The backward computation due to our FedSCAl framework remains almost the same. We save computation by storing pseudo-labels generated by the first forward pass through the global model parameters for the weakly augmented samples. Our method communicates only the model parameters and incurs no additional communication cost.

\begin{table}[h]
\centering
\caption{Training time comparison of different methods.}
\begin{tabular}{l c}
\toprule
\textbf{Method} & \textbf{Time (sec)} \\
\midrule
FedLoA   & 156.93 \\
LADD     & 114.90 \\
FedProx  & 158.28 \\
FedWCA   & 163.83 \\
FedSCAL  & 170.08 \\
\bottomrule
\end{tabular}
\label{tab:time_comparison}
\end{table}

The Table \ref{tab:time_comparison} reports the end-to-end training time (in seconds) for different federated learning baselines for single communication round. Among all methods, FedSCAL, our proposed approach, shows a moderate increase in computation time compared to others. This is expected, as FedSCAL incorporates both local and global consistency mechanisms. Despite this slight overhead, FedSCAL delivers significantly better performance, demonstrating that the additional computation is minimal relative to its overall gains.

\section{On the use of Local and Global models}

As demonstrated in our experiments, both the local and global consistency losses 
are essential to the effectiveness of the SCAL mechanism. We now formalize why this 
is the case. Let $p_{l,k}^i$ denote the probability that the \emph{local} model of 
client $k$ incorrectly predicts the weakly augmented sample $i$. Similarly, let 
$p_{g}^i$ denote the error probability of the \emph{global} model on the same 
sample. The consistency loss for sample $i$ receives incorrect supervision only 
when both models are wrong, which occurs with probability $p_{l,k}^i\, p_{g}^i$. 
Thus, the probability of receiving correct supervision is
\[
1 - p_{l,k}^i\, p_{g}^i.
\]

Notice that
\[
1 - p_{l,k}^i\, p_{g}^i > 1 - p_{l,k}^i
\qquad\text{and}\qquad
1 - p_{l,k}^i\, p_{g}^i > 1 - p_g^i,
\]
since $p_{l,k}^i\, p_{g}^i < p_{l,k}^i$ and $p_{l,k}^i\, p_{g}^i < p_g^i$. 
In other words, combining both local and global predictions strictly increases 
the probability of obtaining correct supervision compared to relying on either 
model alone. This explains why the joint use of local and global consistency 
losses yields more reliable pseudo-labels and contributes significantly to the 
success of SCAL.

\end{document}